%% file: acl_latex.tex
\documentclass[11pt]{article}

\usepackage[preprint]{acl}
\usepackage{times}
\usepackage{latexsym}
\usepackage{wrapfig}
\usepackage{microtype}
\usepackage{hyperref}
\usepackage{url}
\usepackage{booktabs}
\usepackage{graphicx}
\usepackage[table]{xcolor}
\usepackage{colortbl}
\usepackage[most]{tcolorbox}
\usepackage{listings}
\definecolor{listinggray}{gray}{0.9} 
\definecolor{headergreen}{RGB}{119, 171, 90} 
\usepackage{longtable}
\usepackage{array}
\definecolor{barred}{HTML}{5B8DEF}
\definecolor{barblue}{HTML}{EF5B5B}

\usepackage[T1]{fontenc}
\usepackage[utf8]{inputenc} 
\usepackage{caption}
\usepackage{amsfonts}

\usepackage[T1]{fontenc}

\usepackage[utf8]{inputenc}

\usepackage{microtype}

\usepackage{inconsolata}
\usepackage{fontawesome5}
\usepackage{graphicx}
\usepackage{subcaption, amssymb}
\usepackage{tabularx,booktabs,xcolor,colortbl}
\definecolor{traceframe}{HTML}{1F4E79}
\definecolor{tracequote}{HTML}{F5F5F5}
\definecolor{tracecheck}{HTML}{2E7D32}

\title{\textcolor{orange!60!yellow}{\faSun}~\ours{}: Eliciting General Reasoning in LLMs with Reinforcement Learning on Natural Instructions}

\author{Ashima Suvarna, Kendrick Phan, Mehrab Beikzadeh, Hritik Bansal, \textbf{Saadia Gabriel} \\
University of California, Los Angeles \\
\faGithub~ \texttt{\href{https://github.com/asuvarna31/supernova}{github.com/asuvarna31/supernova}}\\
}

\newcommand{\ours}{\textsc{SuperNova}}
\newcommand{\eight}{\text{pass@8}}
\newcommand{\one}{\text{pass@1}}

\begin{document}
\maketitle
\begin{abstract}
\input{sections/00-abstract}

\end{abstract}

\input{sections/01-introduction}

\input{sections/02-background}
\input{sections/04-method}

\input{sections/05-setup}

\input{sections/06-curation_results}

\input{sections/07-results}

\input{sections/08-conclusion}

\bibliography{custom}

\appendix
\clearpage
\input{sections/09-app}

\end{document}

%% file: sections/00-abstract.tex
Reinforcement Learning with Verifiable Rewards (RLVR) has substantially improved reasoning in formal domains such as mathematics and code, but extending these gains beyond STEM remains challenging. Extending RLVR beyond STEM is fundamentally constrained by the lack of high-quality verifiable training data. In this work, we introduce \ours{}, a framework for curating RLVR data from natural instruction datasets, which are a rich source of expert-annotated data but are underexplored for RLVR training. Through $100+$ controlled RL experiments, we systematically study how to utilize these dataset for RLVR and how data curation decisions affect downstream reasoning performance . In particular, we investigate three data designs: (a) source task selection, (b) task mixing, and (c) synthetic interventions. Our analysis reveals that source task selection has a significant impact on downstream reasoning performance. Moreover, selecting tasks based on their performance for individual target tasks outperforms strategies based on overall average performance and synthetic interventions do not improve reasoning. Guided by these insights, we construct \ours{}, a high-quality RLVR dataset of $25$K instances curated from natural instruction datasets. We show that training Qwen3-0.6B on \ours{} outperforms the base Qwen3-0.6B, yielding a relative gain of $64.4$pp on BigBench Extra Hard (BBEH), a challenging benchmark comprising 23 complex reasoning tasks. Importantly, we find that gains from \ours{} generalize to unseen benchmarks, larger model scales, and newer model families. Overall, our findings provide practical insights for curating human-annotated resources to extend RLVR to general reasoning.

\begin{figure*}[h]
    \centering
    \begin{subfigure}{0.48\linewidth}
        \centering
        \includegraphics[width=\linewidth]{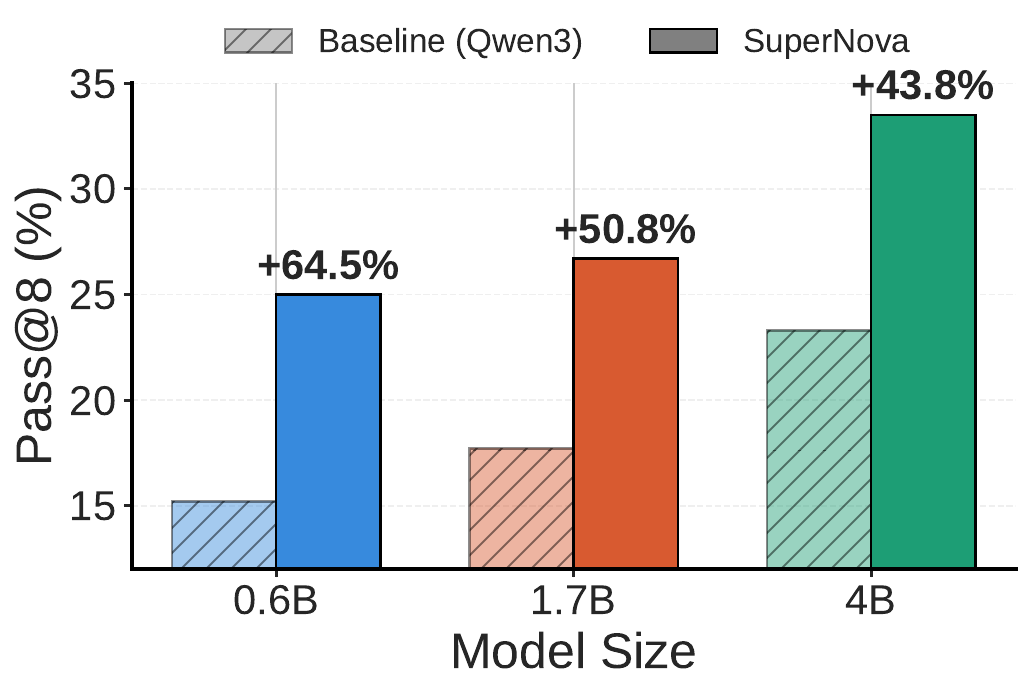}
        \caption{}
        \label{fig:pull-a}
    \end{subfigure}
    \hfill
    \begin{subfigure}{0.48\linewidth}
        \centering
        \includegraphics[width=\linewidth]{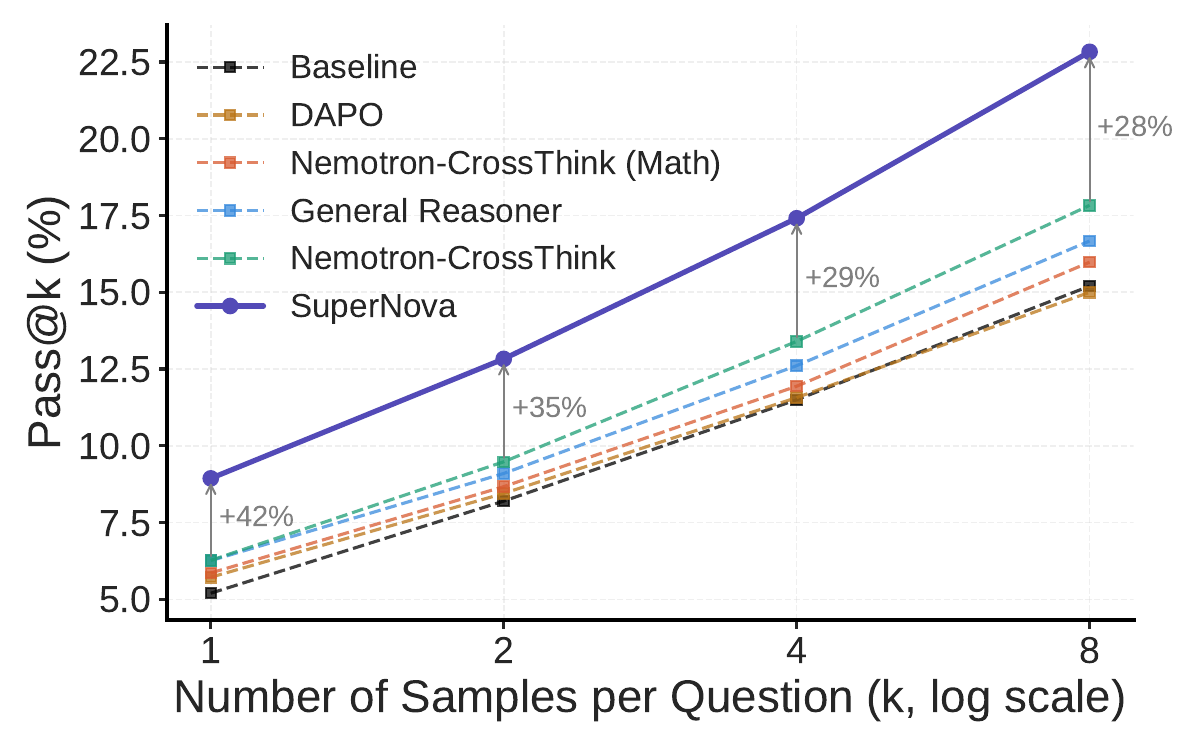}
        \caption{}
        \label{fig:pull-b}
    \end{subfigure}
    
    \caption{\textbf{\ours{} shows strong reasoning performance.} Training with our curated \ours{} data leads to consistent pass@k improvements on a challenging benchmark, BBEH-test. (a) \ours{} shows consistent performance gains across various model sizes from the Qwen3 family. (b) \ours{} shows superior performance to existing reasoning datasets across varying values of k under compute-matched comparisons. }
    \label{fig:abstract}
\end{figure*}

%% file: sections/01-introduction.tex
\section{Introduction}
 Large language models (LLMs) have shown remarkable progress in reasoning capabilities for formal domains such as mathematics and code~\citep{guo2025deepseek,lambert2024tulu,guha2025openthoughts,ma2025general,zeng2025simplerl,hu2025open,chen2025acereason}. A vast majority of these advances leverage reinforcement learning with verifiable rewards (RLVR) which relies on the ability to verify model outputs against a ground-truth final answer \citep{guo2025deepseek}. The success of RLVR in STEM domains is closely tied to the abundance of high-quality verifiable data, such as MATH~\citep{hendrycks2021measuringmathematicalproblemsolving}, competitions (e.g., CodeForces, Art of Problem Solving) and forums (e.g., StackOverflow). Together, these resources have enabled the rapid adoption of RLVR for mathematical and code reasoning~\citep{chen2025acereason,yu2025dapo,akter2026nemotron,hu2025open}. 

 However, reasoning in real-world settings extends beyond STEM problem solving and requires a broader spectrum of capabilities, including temporal reasoning, causal inference, and logical deduction~\citep{newell1972human,johnson2010mental,griffiths2020understanding}. Yet recent work shows that training models on mathematical and coding data does not necessarily improve general reasoning capabilities~\citep{bhaskar2025languagemodelsthinkchat,huan2025doesmathreasoningimprove,cheng2026revisiting}. In particular, models trained on high-quality mathematical and coding reasoning data often achieve substantial gains on formal reasoning benchmarks while simultaneously degrading performance on more general reasoning tasks. For example, OpenReasoner-7B~\citep{hu2025open} and OpenThinker-7B~\citep{guha2025openthoughts} improve over their base models by more than $50\%$ on challenging mathematical benchmarks such as AIME24~\citep{aime24}, yet suffer performance drops of up to $8\%$ on complex reasoning tasks in BBEH~\citep{kazemi2025big}.


 A key bottleneck in extending RLVR beyond STEM domains is the lack of high-quality verifiable training data that spans diverse reasoning skills. Prior work addresses this challenge through domain-specific reward designs and extensive data filtering pipelines to curate reasoning data spanning diverse domains~\citep{cheng2026revisiting,ma2025general,akter2026nemotron}. However, these approaches often depend on noisy web-scale corpora~\citep{ma2025general} or specialized datasets tailored to particular domains~\citep{cheng2026revisiting}, limiting their applicability to understudied reasoning tasks. Moreover, collecting human-annotated data for RLVR is expensive and labor-intensive.

 In contrast, large amounts of high-quality human-annotated data already exist in instruction-following datasets. Resources such as SuperNI~\citep{wang2022super} and FLAN~\citep{wei2021finetuned} contain thousands of expert-curated tasks spanning event understanding, question generation, and other reasoning-intensive settings (Appendix Table~\ref{app_tab:superni}). However, these datasets cannot be directly used for RLVR because (a) many tasks are open-ended and lack reliable verification signals, (b) not all tasks elicit strong reasoning behavior, and (c) the principles underlying effective RLVR data curation remain underexplored beyond STEM domains. 
 
\begin{figure*}[t]
    \centering
    \includegraphics[width=\linewidth]{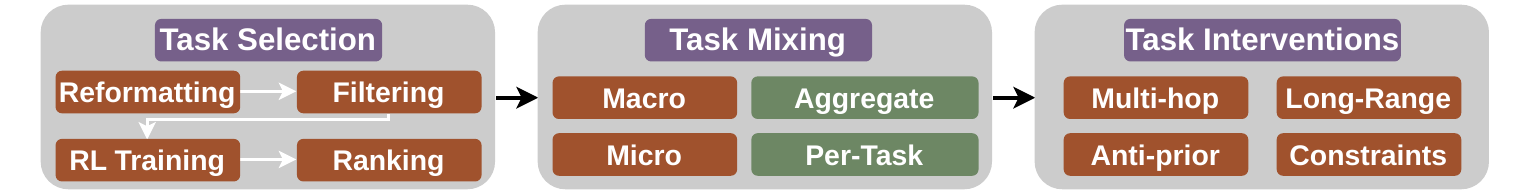}
    \caption{\textbf{\ours{} Framework: } We study the impact of various data design choices for RLVR data curation from natural instruction datasets. First, we study the impact of task selection on downstream reasoning performance. Then, we explore strategies to mix diverse tasks in source data. Finally, we examine whether synthetic data interventions can enhance data quality and improve downstream reasoning.}
    \label{fig:supernova_framework}
\end{figure*}

In this work, we introduce \ours{}, a multi-stage pipeline for curating high-quality RLVR data from natural instruction datasets (Figure~\ref{fig:supernova_framework}). We conduct $100+$ compute-matched RL experiments to systematically study the principles underlying RLVR data curation. We explore three key design choices: (a) source task selection, (b) task mixing strategies, and (c) synthetic interventions to enhance task quality. In particular, we find that (a) task selection has a large impact on downstream reasoning performance, (b) our novel mixing strategy: micro-mixing yields further gains, and (c) synthetic interventions on tasks do not improve reasoning performance. Finally, based on these insights we construct \ours{}, a curated corpus of $25$K verifiable training instances spanning diverse reasoning types for RLVR (Figure~\ref{fig:data_stats}). Training Qwen3 models across scales (0.6B–4B) on \ours{} yields substantial gains on BBEH~\cite{kazemi2025big}, achieving relative gains up to $64.4$pp, while outperforming existing reasoning datasets such as Nemotron-Crossthink~\cite{akter2026nemotron} by $42$pp at \one{} (Figure~\ref{fig:abstract}). Models trained on \ours{} also generalize strongly to challenging reasoning benchmarks including BBH~\citep{suzgun2023challenging}, MMLU-Pro~\citep{wang2024mmlu}, and Zebralogic~\citep{lin2025zebralogic}, while exhibiting cross-model transfer across both different model families. Moreover, these gains remain consistent at larger values of $k$, highlighting the importance of principled RLVR data curation for improving reasoning capabilities in LLMs.

In summary, our contributions are as follows:

\begin{itemize}
    \item \textbf{Natural Instructions can elicit reasoning.} We demonstrate that natural instruction datasets are a rich source of human-annotated data which can be used for RL training to improve complex reasoning. 
    \item \textbf{\ours{} Framework.} We introduce a framework to curate high-quality RLVR data beyond STEM. Through our controlled experiments, we study the impact of diverse data designs including (a) task selection, (b) task mixing, and (c) synthetic interventions. 
    \item \textbf{Generalization of \ours{}.} We curate a high-quality RLVR dataset of $25$K instances. Importantly, we show that gains from \ours{} generalize to evaluation benchmarks that were unseen during data curation. Specifically, our data curated from a small Qwen3-0.6B model generalizes to larger model sizes and newer model families.
\end{itemize}

%% file: sections/02-background.tex
\section{Preliminaries}
\label{sec:background}

\paragraph{Reinforcement Learning with Verifiable Rewards (RLVR).} RLVR is widely adopted for training LLMs for reasoning in domains that rely on automatically verifiable ground truth such as mathematics and code. Given an input-target pair $(q,t)$, RLVR samples G rollouts ${o_i}_{i=1}^{G}$ from a behavior policy $\pi_{\theta_{\text{old}}}$ and optimizes the GRPO~\citep{shao2024deepseekmath} objective:
 
\begin{equation}
\resizebox{\columnwidth}{!}{$
\displaystyle
\mathcal{J}_{\text{GRPO}}(\theta) = \mathbb{E}_{\substack{(q,t) \sim \mathcal{D} \\ \{o_i\}_{i=1}^{G} \sim \pi_{\theta_{\text{old}}}(\cdot \mid q)}} \Bigg[ \frac{1}{G} \sum_{i=1}^{G} \min\Big( \rho_i(\theta)\, \hat{A}_i,\; \mathrm{clip}\!\left(\rho_i(\theta), 1{-}\epsilon, 1{+}\epsilon\right) \hat{A}_i \Big) \Bigg]
$}
\label{eq:grpo}
\end{equation}
 
where $\rho_i(\theta) = \frac{\pi_\theta(o_i \mid q)}{\pi_{\theta_{\text{old}}}(o_i \mid q)}$ is the importance sampling ratio. The group-centered advantage $\hat{A}_i$ for each output is computed as $\hat{A}_i = r_i - \frac{1}{G}\sum_{j=1}^{G} r_j$ where $r_i = r(o_i,q)$, the computed reward. Following~\cite{yu2025dapo}, we skip the KL penalty to improve training efficiency in our experiments.

\paragraph{Task-Specific Instruction Datasets.} Instruction-tuning datasets such as SuperNI~\citep{wang2022super}, and Flan-Collection~\citep{wei2021finetuned} are a collection of well-structured, distinct tasks spanning diverse reasoning abilities. These datasets are constructed from high-quality human supervision including task definitions, instructions and ground-truth annotations. We observe that these large instruction-tuning datasets often encode reasoning structures that are not explicitly annotated but can be inferred from the examples and task structure. Consider an instruction dataset $D = \{D_1, D_2, D_3...D_K\}$ comprising K tasks where each subset $D_k$ is a well-defined task targeting a particular skill. 


\paragraph{Problem Setup.} In this work, we focus on the curation of high-quality training data to enable strong general reasoning capabilities via reinforcement learning. Given a pool of candidate datasets $D = \{D_1, D_2, \ldots, D_K\}$, a model $M$, and a training algorithm $A$, we seek a subset of tasks $S \subseteq D$ that maximizes downstream performance after training. Following the data curation formulations propose for SFT in math reasoning~\citep{guha2025openthoughts} and multimodal reasoning~\citep{bansal2025honeybee}, we define our objective as:
\begin{equation}
    S^*=argmax_{S\subseteq D} \Phi(A(M,S),V)
\end{equation}
where $A(M,S)$ denotes the model after applying $A$ to $M$ on the selected subset $S$, V is the validation set and $\phi$ measures downstream performance on $V$.

\begin{figure}
    \centering
    \includegraphics[width=\linewidth]{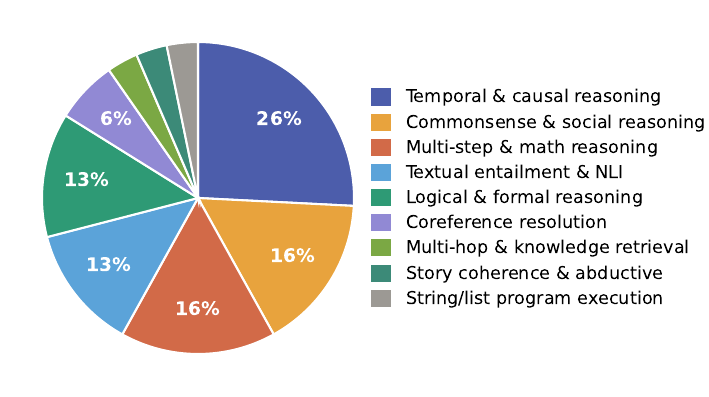}
    \caption{\textbf{\ours{} Distribution.} \ours{} comprises of 25K training samples that target 9 diverse reasoning types.}
    \label{fig:data_stats}
\end{figure}

%% file: sections/04-method.tex
\section{\ours}

We outline our \ours{} framework (Figure~\ref{fig:supernova_framework}), which consists of multiple stages:
(a) task selection, which assesses the impact of source task choice (\S~\ref{sec:task_sourcing}); (b) mixing, which identifies the best strategy to mix the diverse tasks (\S~\ref{sec:data_mixing}); and (c) data interventions, which aim to enhance the quality of our data (\S~\ref{sec:interventions}). Finally, we combine the best performing strategy from each step and curate \ours{}, comprising of $25K$ verifiable training samples spanning 9 diverse reasoning types as shown in Figure~\ref{fig:data_stats}. \ours{} consists of 31 tasks selected from SuperNI with $25\%$ of the tasks targeting temporal and causal reasoning.\footnote{Most SuperNI tasks target various reasoning types, we consider the primary reasoning type identified by Claude-Opus-4.6 based on the task description. More details in App.~\S~\ref{app:task_analysis}.} We provide qualitative examples from \ours{} in Appendix Table 4. 

\subsection{Task Selection}
\label{sec:task_sourcing}
\textbf{Extracting reasoning data from instructions.} The quality of the input and the coverage of reasoning types is critical for determining the reasoning skills imparted to the LLM. For example, a LLM exposed to temporal graphs will excel in downstream temporal understanding tasks~\citep{xiong2024large}. In this work, we leverage instruction-tuning data $D$ to source diverse tasks $D_k$ for general reasoning. Since, instructions are formatted for supervised-finetuning they are not directly usable for RLVR as they may incorporate hard to verify ground-truth. Thus, for every instruction $p$ in $D_k$, we \textbf{reformat the instruction} to a verifiable question $q$. To further identify the most effective data from $D$, we sample 8 rollouts from model $M$ for each $q$ and compute the \textbf{per-question win-rate}. Finally, we remove all questions which are too easy for the model (win-rate=1) or too challenging (win-rate=0). 

\textbf{Task Ranking.} For each task $D_k$, we define a task-utility score $u_k \in R$ that indicates how effective $D_k$
is for RLVR training. Then, we rank the $K$ tasks according to their utility scores, producing a ranking: $u_{D_1} > u_{D_2} > u_{D_3}> \cdots >u_{D_K)}$. The task utility scores enable us to select high-quality tasks while downweighting poor and irrelevant tasks.  We explore various approaches to compute task-utility: (a) we compute the semantic and lexical similarity between the task questions and the questions from our validation benchmark $V$; (b) we compute the difficulty of the task based on the average win-rate of the task under model $M$; and (c) we train model $M$ on $D_k$ and evaluate performance on $V$.

\subsection{Mixing}
\label{sec:data_mixing}
After obtaining high-quality tasks, we determine how to combine them to construct an effective training mixture. Mixing strategy is a key design choice in data curation and prior work in LLM reasoning have shown to yield superior datasets by mixing subsets from various sources. Consider the K tasks from \S~\ref{sec:task_sourcing} and number of tasks to be mixed $N \in \{1,2,4,8,16\}$, we want to determine the optimal value of N under two mixing strategies:
\begin{itemize}
    \item \textbf{Macro Mixing:} Consider the ranking from \S~\ref{sec:task_sourcing}: $u_{D_1} > u_{D_2} > u_{D_3} ..>u_{D_K}$ where $u_{D_k}$ is the macro average of model performance on $V_{BBEH}$. We select the top-ranked N tasks $u_{D_1} > u_{D_2} > u_{D_3}> \cdots>u_{D_N}$ for our mixture. 
    \item \textbf{Micro Mixing:} Here, we leverage the sub-tasks of our $V$ and produce a ranking for each sub-task $V_i$. Specifically, we define $u_k^{(i)}$ as the performance of model $M$ trained on $D_k$ and evaluated on sub-task $V_i$, yielding a per-sub-task ranking: $u_{D_1}^{(i)} > u_{D_2}^{(i)} > \cdots > u_{D_K}^{(i)}$ for each $V_i \in V$. We then select the top-ranked $N$ tasks per sub-task and take the unique set of selected tasks for our mixture.
\end{itemize}

\subsection{Data Interventions}
\label{sec:interventions}
Starting from the best mixture from~\S\ref{sec:data_mixing}, we assess whether we can enhance the data quality through targeted data interventions. RLVR datasets primarily focus on the questions since interventions on the target may hinder the verifiability of the answer. Thus, we apply a set of interventions to transform the difficulty of the questions while preserving the target answer. These interventions aim to increase the difficulty of the questions by introducing diverse reasoning types such very long-context dependency, information that prompts model to go against a strong prior or needle in haystack. Let $D_{\text{base}} = \{(q, t)\}$ be the base data with original question-target pairs. We apply an intervention $I$ that transforms each question while preserving the target, producing $D' = \{(I(q), t)\}$. We provide the implementation details of applying these interventions in Appendix \S~\ref{app:interventions}.

%% file: sections/05-setup.tex

\begin{figure*}[t]
    \centering
    \includegraphics[width=0.9\linewidth]{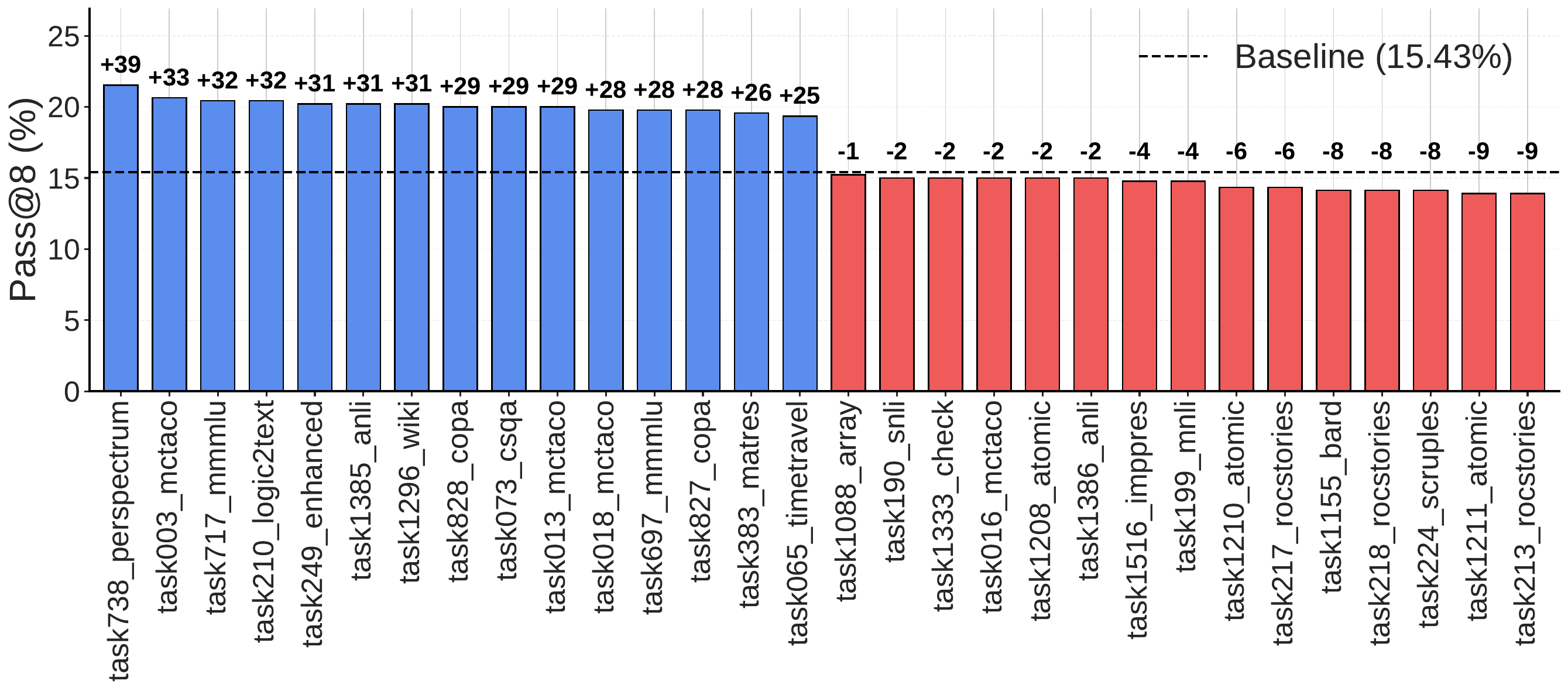}
    \caption{\textbf{Impact of Task Selection.} We train the baseline (Qwen3-0.6B) on each task individually under compute-matched settings.  We report relative pass@8 gains on BBEH-mini for each task and highlight the tasks that \colorbox{barred!30!white}{improve} and \colorbox{barblue!30!white}{degrade} the baseline.}
    \label{fig:task_ranking}
\end{figure*}

\section{Experimental Setup}

\paragraph{Training Data.} We use SuperNI~\citep{wang2022super} as our data source. It consists of 1600 tasks spanning various tasks types such as question answering, question generation and commonsense reasoning. Each task consists of the task description and the instruction-response pair, annotated by experts. We sample a candidate pool of 83 tasks for our experiments.\footnote{The candidate pool allows us to implement \ours{} on tractable compute. Ideally, \ours{} can be applied to all tasks from SuperNI. Additional details in Appendix~\S\ref{app:setup}.} 

\paragraph{Training.} We train models from Qwen3~\cite{yang2025qwen3} family (0.6B, 1.7B and 4B) with GRPO~\citep{shao2024deepseekmath} for all our experiments (\S~\ref{sec:background}). We adopt binary rewards for training: 1 if the output is judged correct under rule-based verification and 0 otherwise. The rule-based verifier extracts the final answer, normalizes the output and applies string matching. For our data curation experiments, we use Qwen3-0.6B for faster training iterations. All our data curation experiments were run for 250 RL steps. Finally, we train the \ours{} models for 5000 RL steps. We present more details about the training setup in Appendix~\S\ref{app:setup}. 

\paragraph{Evaluation.} We evaluate our models on various benchmarks that target diverse reasoning capabilities. For our data curation experiments, we choose BBEH-mini as our validation benchmark. BBEH-mini is a small subset of BBEH~\citep{kazemi2025big} that comprises of 23 challenging tasks that target diverse reasoning capabilities. We use the remaining BBEH samples, which are not included in BBEH-mini as the unseen test set, BBEH-test. Thus, downstream reasoning performance is aggregated over all 23 tasks. After curating \ours{}, we evaluate our models on 4 additional \textit{unseen} benchmarks including BBH~\citep{suzgun2023challenging}, Zebralogic~\citep{lin2025zebralogic}, MMLU-Pro~\citep{wang2024mmlu} and MATH500~\citep{lightman2023let}. To ensure consistency, we use an identical prompt across all evaluations that encourages the model to think before answering, provided in Appendix~\S\ref{app:eval_details}. 

\paragraph{Evaluation Metric.}
We adopt $\text{pass@k}$ as our evaluation metric, which is well-suited for evaluating RL-trained models~\citep{chen2021evaluating,yue2025limit-of-rlvr}. As shown in 
Appendix \S~\ref{app:passk_correl}, we find that $\text{pass@8}$ provides 2.5 times greater discriminability than $\text{pass@1}$ ($\sigma$: 0.76 
$\rightarrow$ 1.92). We therefore utilize $\text{pass@8}$ for our data curation experiments. 

\begin{figure*}[t]
    \centering
    \includegraphics[width=\linewidth]{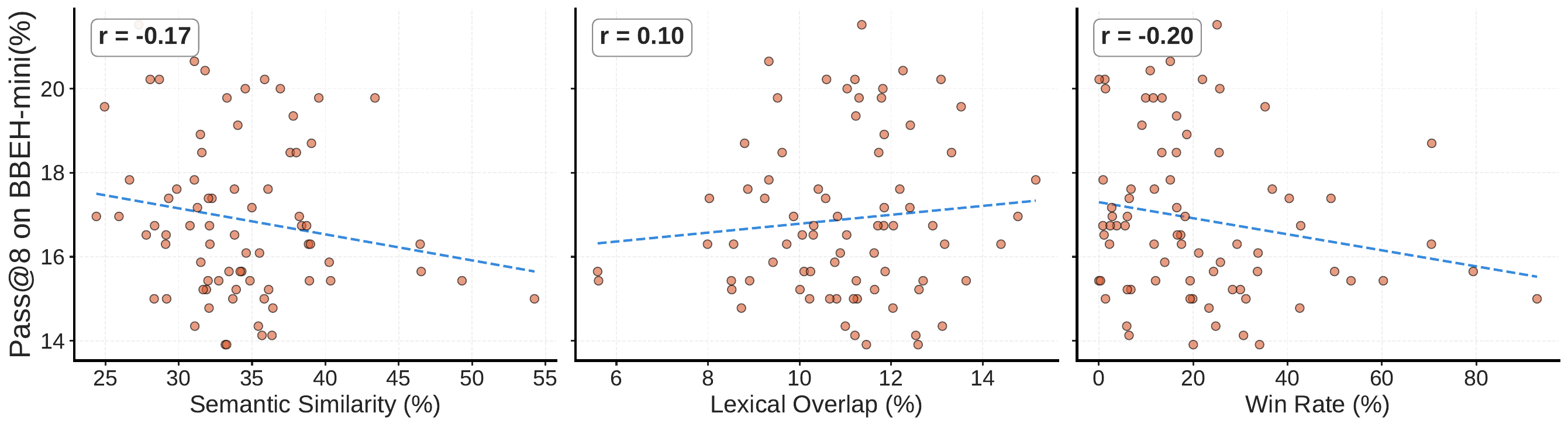}
    \caption{\textbf{(Left)} Correlation between semantic similarity and task performance. \textbf{(Middle)} Correlation between lexical similarity and task performance. \textbf{(Right)} Correlation between win rate and downstream reasoning performance. }
    \label{fig:correl}
\end{figure*}

\paragraph{Baselines.}
To further compare the quality of \ours{} with other reasoning datasets, we train Qwen3-0.6B under a compute-matched setup using three baseline datasets including Nemotron-CrossThink~\citep{akter2026nemotron}, which targets diverse domains and question-answer formats; General-Reasoner~\citep{ma2025general}, which curates reasoning data across diverse STEM-focused domains; and DAPO~\citep{yu2025dapo}, a high-quality math reasoning dataset sourced from competition websites. We provide additional details in Appendix~\S\ref{app:baseline_data}. We evaluate several models as baselines for our experiments. 
\textbf{(1) Qwen3}: included to measure the gains obtained from training on \ours{}.
\textbf{(2) OpenThinker3-7B}~\citep{guha2025openthoughts}: a strong math reasoning model supervised finetuned on large math corpus.
\textbf{(3) OpenReasoner-Nemotron-7B}~\citep{ahmad2025opencodereasoning}: a strong reasoning model. \textbf{(4) Olmo3-7B-Think}~\citep{olmo2025olmo}: a state-of-art reasoning model that has strong performance across various domains.

%% file: sections/06-curation_results.tex
\section{Experiments}

\paragraph{Impact of Task Selection.}
We train \texttt{Qwen3-0.6B} on each task and report model performance on BBEH-mini in Figure~\ref{fig:task_ranking}. Our experiments show that task selection has a substantial impact on downstream reasoning performance (Figure~\ref{fig:task_ranking}). Specifically, we observe a 7.6 percentage point (pp) gap between the lowest-performing task (\texttt{task213-rocstories}, $13.9\%$) and the highest-performing task (\texttt{task738-perspectrum}, $21.5\%$). Notably, several tasks degrade performance relative to the baseline, underscoring that not all tasks are beneficial for improving reasoning under RLVR, and task selection based on task utility is critical for training strong reasoning models. Furthermore, we find that tasks involving multi-hop reasoning yield the largest gains over the baseline model (Appendix \S~\ref{app:task_analysis}). 

\paragraph{Task Utility Ranking.}
 We find that semantic similarity and lexical similarity between the tasks and validation benchmark are poor predictors of task utility for RLVR. As shown in Fig.~\ref{fig:correl}, both measures exhibit weak correlation with model performance on BBEH-mini. These approaches are attractive because they are cheap, fast to implement, and model-agnostic. However, our findings suggest that surface similarity is insufficient for task selection for RLVR. We also investigate whether task difficulty, measured by the average win-rate of the base model, predicts downstream reasoning performance in Fig.~\ref{fig:correl}. Similar to surface similarity, we observe only a weak correlation between task difficulty and model performance on BBEH-mini, indicating that base-model difficulty is also a poor predictor of task utility for RLVR. Overall, our findings underscore that effective task selection is grounded in compute-matched RL training, and simpler proxies are insufficient for predicting downstream performance.  

\input{tables/split_table}
\input{tables/datasets_ood}

\paragraph{Impact of Task Mixing.}
\label{sec:mixing_results}
We present the results of two mixing strategies: Macro Mixing and Micro Mixing in Table~\ref{tab:mixing}.  Across both strategies, we find that mixing top 4, 8 or 16 tasks did not yield better results than top 1 or 2 tasks at both \one{} and \eight{}. Thus, mixing strategies are highly dependent on the training data distribution and mixing more data sources does not always yield high reasoning performance. Furthermore, we find that micro-mixing with top-2 achieves the highest \eight{} scores of $22.8\%$ our experiments. These results indicate that selecting top-ranked tasks per sub-task (Micro Mixing) yields better performance than selecting tasks based on overall ranking at compute-matched settings.

\paragraph{Impact of Data Interventions.}
We apply several data intervention strategies to the best-performing dataset from \S\ref{sec:mixing_results} (Micro-Top2) and report results on BBEH-mini in Appendix Table~\ref{tab:interventions_results}. Surprisingly, none of the interventions improve over the original data. While Going Against Prior achieves the highest performance among the interventions ($22.6\%$), is comparable to but does not exceed Micro Top2. Thus, such synthetically generated interventions on the question are not effective in improving downstream reasoning performance with RLVR. 
\begin{figure}
    \centering
    \includegraphics[width=\linewidth]{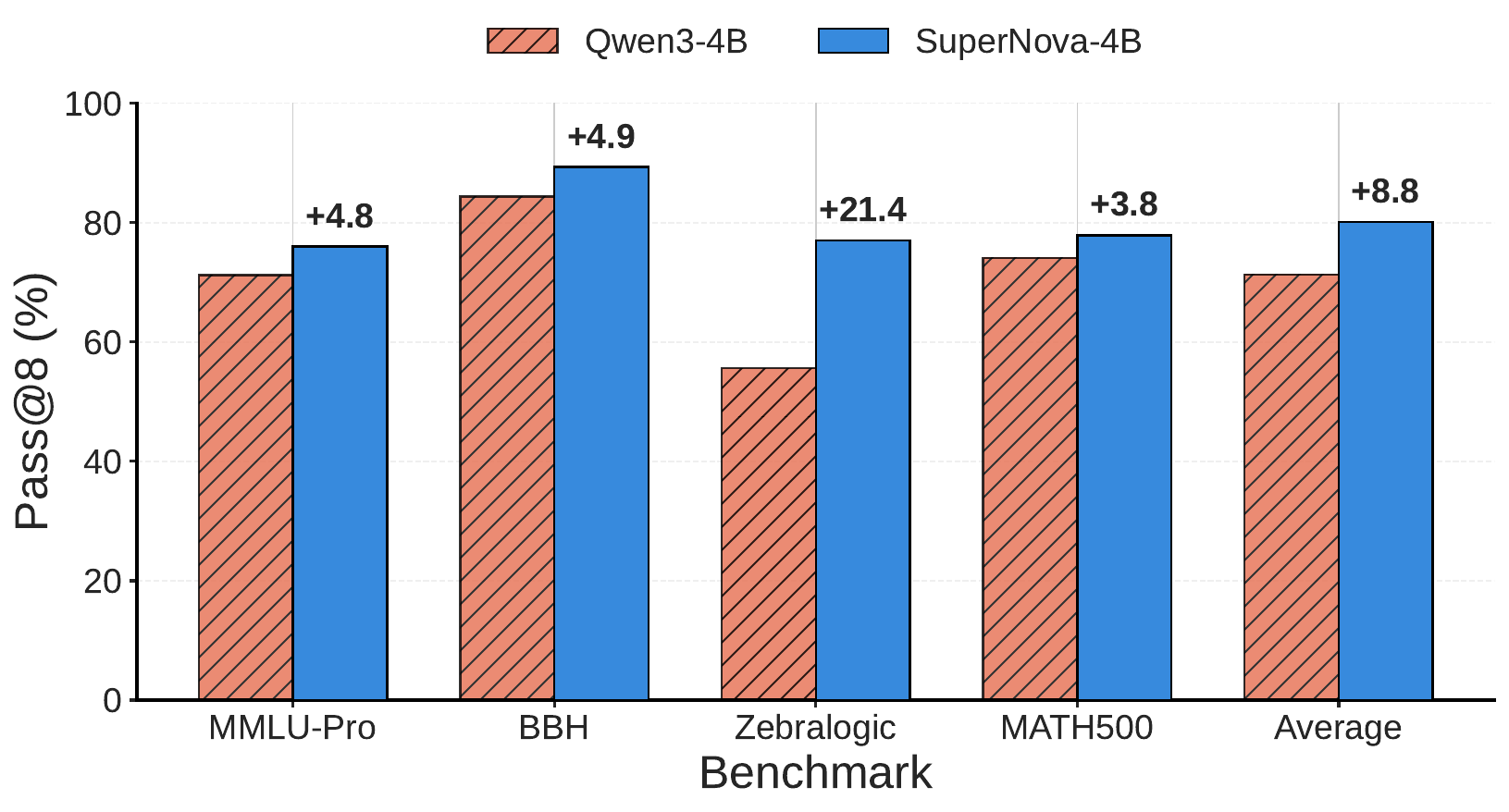}
    \caption{\textbf{\ours{} generalizes to OOD Benchmarks.} We report \eight{} across four reasoning benchmarks for Qwen3-4B and \ours{}-4B. }
    \label{fig:ood_fig}
\end{figure}

%% file: tables/split_table.tex
\begin{table}[t]
\footnotesize
  \begin{tabular}{lccccc}
    \toprule
    & \textbf{Top 1} & \textbf{Top 2} & \textbf{Top 4} & \textbf{Top 8} & \textbf{Top 16} \\
    \midrule
    \multicolumn{6}{c}{\textit{Micro Mixing}}\\
    \midrule
    pass@1 & 7.5 & \textbf{8.9} & 7.5 & 7.6 & 7.5 \\
    pass@8 & 18.3 & \textbf{22.8} & 18.7 & 18.0 & 20.2 \\
    \midrule
    \multicolumn{6}{c}{\textit{Macro Mixing}}\\
    \midrule
    pass@1 & 7.6 & 8.2 & 6.6 & 6.4 & 7.5 \\
    pass@8 & 21.5 & 21.7 & 17.4 & 17.0 & 18.3 \\
    \bottomrule
  \end{tabular}
  \vspace{0.6em}
  \caption{\textbf{Impact of mixing.} We mix questions from tasks
    following two strategies: micro mixing and macro mixing. We find
    that micro mixing with top 2 tasks achieves the best performance
    \textbf{(bold)}.}
  \label{tab:mixing}
\end{table}

%% file: tables/datasets_ood.tex
\begin{table*}[]
\centering
\small
\begin{tabular}{lccccc}
\toprule
Model             & MMLU-Pro & BBH    & Zebralogic & MATH500 & Average \\
\midrule
Qwen3-0.6B         & 55.3   & 52.4   & 34.4   & \textbf{71.9}   & 53.5   \\

\quad RL w/ General-Reasoner~\cite{ma2025general}       & 54.4   & 64.3   & 45.4   & 66.3  & 57.6   \\
\quad RL w/ Nemotron-Crossthink~\cite{akter2026nemotron} & 55.7   & 69.9   & 49.1   & 70.0   & 61.2  \\
\rowcolor{blue!8}\quad RL w/ \ours{} & \textbf{56.2}   & \textbf{81.5}   & \textbf{49.4}   & \underline{71.4}   & \textbf{64.6}   \\

\bottomrule
\end{tabular}
\caption{\textbf{\ours{} beats reasoning datasets on reasoning benchmarks}. We report \eight{} across four benchmarks that were \text{unseen} during data curation. We find that LLMs trained on \ours{} show improved \eight{} over the models trained on baseline datasets under compute-matched settings.}
\label{tab:ood_evals}
\end{table*}

%% file: sections/07-results.tex
\section{Training Reasoners with \ours{}}
\label{sec:results}

\input{tables/task_specific_16}
\paragraph{\ours{} demonstrates strong reasoning performance.} We find that training Qwen3-4B on \ours{} outperforms the base Qwen3-4B model with a relative gain of 43.8pp (Figure~\ref{fig:abstract}(a)). Moreover, \ours{} also demonstrates consistent gains over the base model across 0.6B and 1.7B models. We show a reasoning trace produced by \ours{}-4B on a challenging Boardgame-QA example in Appendix Figure 11. The model performs multi-step reasoning by chaining together multiple rules and resolving preference conflicts. This example demonstrates that training on \ours{} enables the model to perform structured reasoning over challenging tasks. 

\paragraph{\ours{} beats SOTA reasoning datasets.} We compare \ours{} against three state-of-the-art reasoning datasets that target diverse reasoning skills. To ensure a fair comparison of data quality, we perform a compute-matched analysis of all datasets (details in Appendix \S~\ref{app:baseline_data}). Results on BBEH-test are shown in Figure~\ref{fig:abstract}(b). We find that \ours{} achieves relative gains of 42pp on $\text{pass@1}$ and 28pp on $\text{pass@8}$ over the strongest baseline, Nemotron-Crossthink. In contrast, both math reasoning datasets, DAPO and Nemotron-Crossthink (Math) show little to no improvement over the baseline. Additionally, we report \eight{} performance of all datasets on OOD benchmarks in Table~\ref{tab:ood_evals}. Models trained on \ours{} outperform those trained on all baseline datasets, achieving 81.5\% on BBH and a 3.4pp improvement on average over Nemotron-Crossthink.

\begin{figure}
    \centering
    \includegraphics[width=0.75\linewidth]{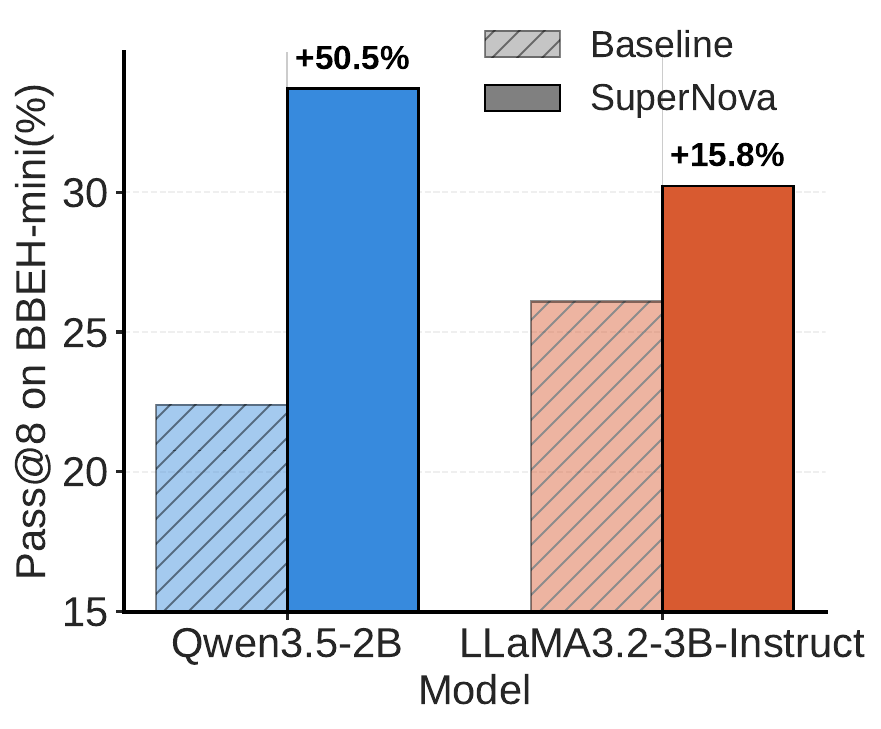}
    \caption{We train Qwen3.5-2B and LLaMA3.2-3B-Instruct with \ours{} and show relative gains over their respective baseline models on 23 complex reasoning tasks. }
    \label{fig:models}
\end{figure}

\begin{figure}
    \centering
    \includegraphics[width=0.75\linewidth]{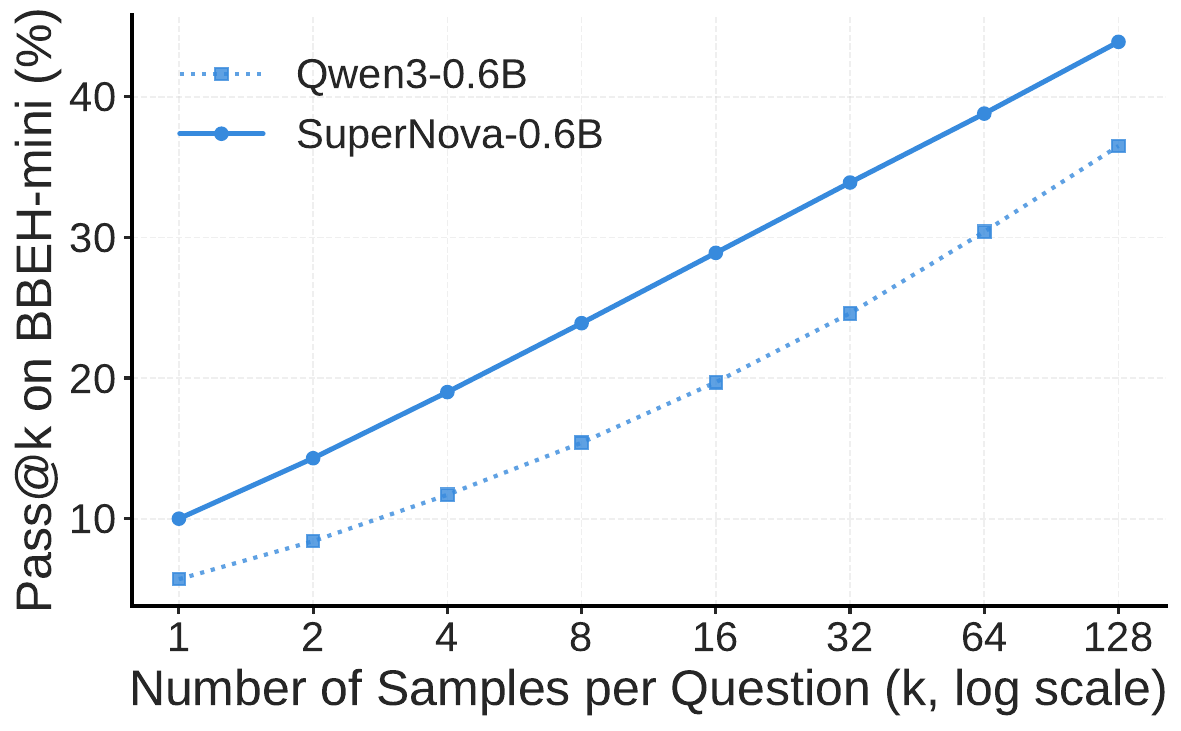}
    \caption{ We show the performance comparison between the baseline model and \ours{}-0.6B by scaling values of k up to 128 on 23 complex reasoning tasks.}
    \label{fig:pass@128}
\end{figure}

\paragraph{\ours{} generalizes to Out-of-Distribution (OOD) Benchmarks.} We evaluate \ours{}-4B on challenging reasoning benchmarks that are unseen during data curation, as shown in Figure~\ref{fig:ood_fig}. Notably, \ours{} achieves substantial gains on Zebralogic, where \ours{}-4B outperforms Qwen3-4B by $21$pp. \ours{}-4B also shows gains of 4.8pp on MMLU-pro and 4.9pp on BBH over the base model. Finally, \ours{}-4B shows gains of 3.8pp on MATH500 despite not being trained on math, indicating that training on \ours{} transfers performance improvements to math reasoning benchmarks.

\paragraph{\ours{} demonstrates reasoning gains over diverse tasks.} To further understand the reasoning performance gains from \ours{}, we report the \eight{} on 13 challenging tasks from BBEH-test on which \ours{}-4B demonstrates substantial gains in Table~\ref{tab:bbeh}. We find that \ours{}-4B consistently outperforms Qwen3-4B across all tasks with an absolute gain of 19pp. \ours{}-4B achieves substantial gains on tasks such as Hyperbaton, Geometric objects and Shuffled objects where Qwen3-4B achieves zero performance. This highlights that \ours{} elicits reasoning behaviors beyond the base model's capabilities on challenging reasoning tasks. Finally, we observe that \ours{}-4B outperforms Qwen3-8B across 12 tasks by 20.3pp despite having half the model capacity. This further highlights that \ours{} is a high-quality dataset that improves parameter efficiency, enabling smaller models to outperform substantially larger ones.

\paragraph{\ours{} gains are consistent at larger values of $k$.}
We analyze whether the performance gains from \ours{} persist at higher values of $k$. As shown in Figure~\ref{fig:pass@128}, \ours{}-0.6B maintains consistent gains over Qwen3-0.6B across all values of $k$ up to 128. This suggests that training on \ours{} expands the model’s exploration space even at large sample sizes, enabling more diverse reasoning behaviors than the baseline.

\paragraph{\ours{} shows cross-model generalization.}
We study how training on \ours{} generalizes across model families (Figure~\ref{fig:models}). In particular, LLaMA3.2-3B-Instruct~\citep{grattafiori2024llama} trained on \ours{} achieves gains of 15.8pp over its baseline. We further observe similar improvements on Qwen3.5-2B~\citep{team2026qwen3}, suggesting that data curation insights derived from earlier-generation models (e.g., Qwen3) transfer to newer-generation models. Overall, the benefits of \ours{} generalize across both model families and generations.

%% file: tables/task_specific_16.tex
\begin{table*}[t]
\centering
\resizebox{\textwidth}{!}{%
\small
\begin{tabular}{lc|ccccccccccccc}
\toprule
Model & \textbf{Avg.} & Brd. & Mov. & Dis. & Bool. & Geo. & T.Ar. & WoL & Word & Shuf. & Zebra & NYCC & M.Ar. & Hyp. \\
\midrule
OLMo-3-7B-Think & 15.1 & 6.1 & \textbf{78.4} & 48.5 & 0 & 0 & 16.2 & 2.4 & 24.4 & 0 & 0 & 20.8 & 0 & 0 \\
OpenReasoning-7B & 9.7 & 4.1 & 19.6 & 60.6 & 0 & 0 & 0 & 2.4 & 9.8 & 0 & 0 & 29.2 & 0 & 0 \\
OpenThinker3-7B & 8.8 & 12.2 & 9.8 & 54.5 & 0 & 0 & 8.1 & 0 & 17.1 & 0 & 0 & 12.5 & 0 & 0 \\
Qwen3-8B & 26.5 & 59.2 & 51 & 57.6 & 5.4 & 2.9 & \textbf{67.6} & 31 & 36.6 & 2.3 & 10 & 20.8 & 0 & 0 \\
\midrule
Qwen3-4B & 25.8 & 65.3 & 64.7 & 51.5 & 5.4 & 0 & 48.6 & 26.2 & 36.6 & 0 & 10 & 25 & 2.2 & 0 \\
\rowcolor{blue!8}\textbf{SuperNova-4B} & \textbf{46.8} & \textbf{71.4} & \underline{65.7} & \textbf{65.2} & \textbf{56.8} & \textbf{54.4} & \underline{51.4} & \textbf{47.6} & \textbf{42.7} & \textbf{41.9} & \textbf{33} & \textbf{30.2} & \textbf{25.6} & \textbf{22.2} \\
\bottomrule
\end{tabular}}
\caption{\textbf{\ours{} models exhibit strong performance across challenging reasoning tasks.} We report the \eight results (\%) on 13 tasks from BBEH-test. Best per column is \textbf{bolded} and second-best in \underline{underlined}.  Avg. is computed across the 13 tasks shown. ( Brd.=Boardgame QA, Mov.=Movie Recommendation, Dis.=Disambiguation QA, Bool.=Boolean Expressions, Geo.=Geometric Shapes, T.Ar.=Time Arithmetic, WoL=Web of Lies, Word=Word Sorting, Shuf.=Shuffled Objects, Zebra=Zebra Puzzles, M.Ar.=Multistep Arithmetic, Hyp.=Hyperbaton.)}
\label{tab:bbeh}
\end{table*}

%% file: sections/08-conclusion.tex
\section{Conclusion}
In this work, we introduce \ours{}, a framework for curating RLVR data from large-scale instruction-tuning datasets to improve the reasoning capabilities of LLMs beyond formal STEM domains. Through controlled RL experiments, we show that effective RLVR data curation depends critically on source task selection and fine-grained task mixing, while synthetic interventions designed to increase reasoning complexity do not reliably improve downstream performance. We further demonstrate that models trained on \ours{} generalize across challenging reasoning benchmarks and transfer across model families and newer model generations. Our findings suggest that existing human-annotated instruction datasets contain rich underutilized signals for RLVR, but unlocking their potential requires principled empirical curation such as \ours{} framework.  Overall, we hope \ours{} provides both a practical resource and a foundation for future work on data curation for RLVR in diverse domains.

\section*{Limitations}
While \ours{} provides key insights and exhibits substantial improvements on academic reasoning benchmarks, several important directions remain for future work. We apply \ours{} to a sample of tasks from SuperNI to demonstrate our key findings and hypothesize that our findings will remain consistent scaling training tasks. Additionally, our work focuses on compute-constrained settings, and future work may explore how these data curation principles scale with substantially larger RL budgets. It is possible that continued RL training might yield even pronounced results~\cite{liu2025synlogicsynthesizingverifiablereasoning}. We use RL-training and empirical validation as measures of task utility in our curation pipeline which are expensive at scale, future work may explore how to apply gradient-based strategies for task selection~\cite{xia2024less} or reward oriented strategies~\cite{wu2024rose} in RLVR. Our evaluation benchmarks give a comprehensive assessment of the impact of principled data curation on reasoning capabilties, however, they do not fully capture real-world setting and decision making capabilities. Future work may apply our insights to other complex reasoning tasks such as multi-agent systems, app development, healthcare, finance. 

\section*{Acknowledgements}
We would like to thank Ayush Agarwal, Genglin Liu, Nishad Singhi and UCLA MARS Lab members for their
insightful discussions and feedback on the draft. 
  
\section*{Ethical Concerns}
\ours{} is sourced from SuperNI tasks which were annotated by humans. Thus, \ours{} inherits the biases from SuperNI and RL training on such data may amplify biases present in the base model or underlying source dataset. In this work, we primarily focus on enhancing the reasoning capabilities of LLMs and do not explicitly address fairness, bias mitigation or value alignment. Future work should systematically evaluate how RL on reaasoning data from natural instruction datasets affects model behavior across sensitive axes such as gender and race. 

\textbf{Disclosure of LLM use in both research and reviewing.} We use ChatGPT and Claude as in our experiments and have provided the relevant prompts. Claude was used in formatting latex tables and code generation for the figures. Finally, we used ChatGPT and Claude to assist with grammar and proof-reading in our paper writing.

%% file: sections/09-app.tex
\input{sections/03-related_work}

\input{tables/supernova_egs}

\section{Pass@k Analysis}
\label{app:passk_correl}
\begin{figure}
    \centering
    \includegraphics[width=\linewidth]{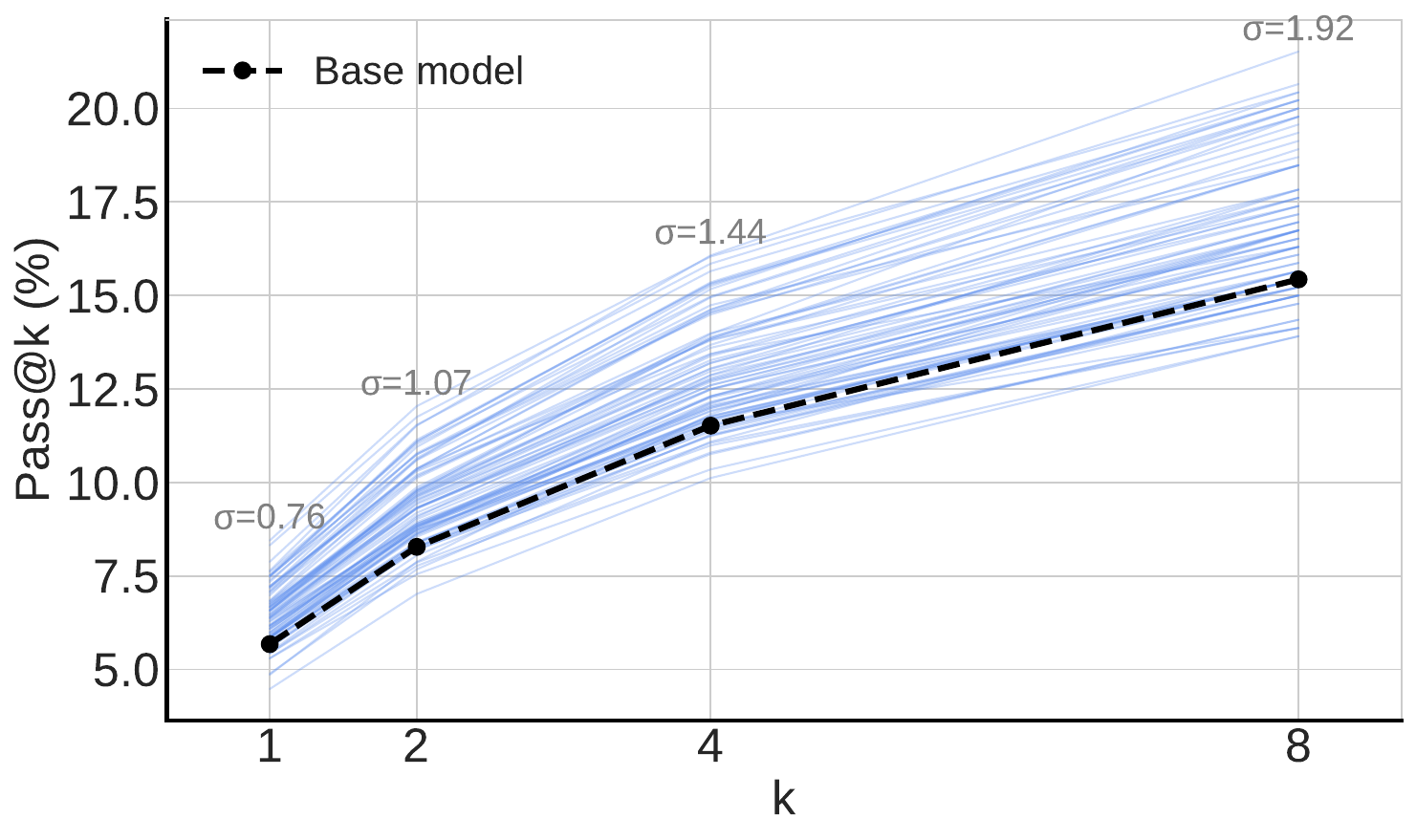}
    \caption{Pass@k accuracy of task-specific models across various values of k.}
    \label{fig:correl_passk}
\end{figure}

Following \cite{chen2021evaluating} and \cite{yue2025limit-of-rlvr}, we analyze the pass@k curves of our task-specific models. Across 80+ RL curves, we observe that the spread and distinguishability of model performance increases at k=8, with maximum overlap at k=1. We show the pass@k curves in Figure~\ref{fig:correl_passk}.

\section{Detailed Experimental Setup}
\label{app:setup}

\subsection{Task Selection}
To ensure we can conduct a controlled study with our limited compute and keep our search space tractable, we prompt an LLM (Claude-opus-4.6) to first classify each task in SuperNI as suitable for reasoning or not, then we randomly sample 83 tasks from all the reasoning suitable tasks to prepare our candidate pool. We would like to note that such capable LLMs are useful for initial task sampling but they do not allow on-policy training and thus exhibit poor correlation with downstream reasoning performance. Ideally, \ours{} can be applied to all tasks from SuperNI and similar instruction-tuning datasets to create large scale RLVR data. We use a minimal binary classification prompt and do not consider the validation benchmark while preparing this candidate pool. This was done to ensure that the task ranking is done purely on task utility scores from~\S 3.1. The prompt follows:

\begin{tcolorbox}[
  colback=gray!8, colframe=gray!40,
  boxsep=2pt, left=4pt, right=4pt, top=2pt, bottom=2pt,
  fontupper=\ttfamily\footnotesize,
  breakable
]
This is an instruction-following task used to train LLMs. Consider the given task description and examples. Now assess the suitability of the task for RL training reasoning models. Think step by step and only respond with yes/no.\\[2pt]
Task ID: \{task\_id\}\\
Task Description: \{description\}\\
Example Input: \{input\}\\
Example Output: \{output\}
\end{tcolorbox}

For reformatting the instruction tuning tasks to verifiable questions, we prompt GPT-5-mini with \ref{lst:reformatting_prompt}. To estimate the quality of the reformatting, we manually inspect 100 samples from 8 tasks and find that GPT-5-mini follows the prompt accurately on 98.4\% of the samples while preserving the ground-truth and original task structure. 

\subsection{Training}
All our experiments were done on 4xH100 gpus. We use the GRPO implementation from TRL\footnote{https://github.com/huggingface/trl} for our training. 
All our data curation experiments utilize Qwen3-0.6B with 500 prompts, learning rate of 1e-6, 8 generations per prompt, batch size of 8, decoding temperature of 0.7 and maximum generation length 4096. We run our training for 250 steps (1 epoch). For our large scale experiments, we run 5000 steps (1 epoch) across 10,000 prompts and use a learning rate of 1e-6 for 0.6B models and 4e-6 for 1.7B and 4B models. 

\subsection{Reward Design}
We adopt binary rewards for all our experiments: 1 if output is judged correct by the reule-based verification and 0 otherwise. The verifier extracts the final answer from the output and matches it against the reference under a normalization-and-fuzzy-match rule: stripping answer-prefix sentinels and LaTeX wrappers (e.g., $\boxed\{\}$, "The answer is:"), lowercasing, and admitting numeric equality, multiple-choice (A)-A equivalence, and list-bracket equivalence. 

\subsection{Evaluation}
\label{app:eval_details}
We use the following benchmarks for our evaluations: 
\begin{itemize}
    \item BBEH~\cite{kazemi2025big}: Comprises of 23 challenging tasks that require diverse reasoning skills such as sarcasm detection, humour detection, constraint satisfaction, boolean algebra, obect ordering, temporal reasoning, commonsense reasoning and logical deduction. Strong reasoning models such as Gemini and O3 achieve 50\% on this benchmark. BBEH was designed to resist saturation and evaluate models on diverse reasoning tasks in a robust manner. 
    \item BBH~\cite{suzgun2023challenging}: Comprises of 23 tasks, similar to BBEH but easier in difficulty. Most modern LLMs have saturated and chain of thought prompting usually achieves great improvements on this benchmark. 
    \item Zebralogic~\cite{lin2025zebralogic}: Comprises of hard and challenging constraint satisfaction logical grid puzzles. 
    \item MMLU-pro~\cite{wang2024mmlu}: tests knowledge intensive reasoning across diverse domains. 
    \item MATH500~\cite{lightman2023let}: A samller subset of the MATH benchmarch that is used to evaluate mathematical reasoning performance of LLMs. 
\end{itemize}
For our evaluations, we use the following prompt across all benchmarks:
\begin{tcolorbox}[
  colback=gray!8, colframe=gray!40,
  boxsep=2pt, left=4pt, right=4pt, top=2pt, bottom=2pt,
  fontupper=\ttfamily\footnotesize,
  breakable
]
Think step by step, and when you are ready to provide the final answer, use the 
prefix "The answer is:" followed by the answer directly,  with no formatting and no 
markup. For instance: "The answer is: 42", or  "The answer is: yes", or "The answer 
is: (a)" For multi-choice questions, provide the letter, e.g. "The answer is: (a)
\end{tcolorbox}
All evaluations were conducted on 1xH100 with a batch size of 8. We use decoding temperature=0.7 with maximum generation length of 4096 across all our experiments. 

\section{Implementation Details of Training Baseline Datasets}
\label{app:baseline_data}
For fair comparison across dataset quality, we train Qwen3-0.6B on the fixed budget of 250 RL steps across 500 prompts and the same learning rate for all datasets. Since Nemotron-Crossthink, Dapo and General-Reasoner are large-datasets, we report their performance as average pass@8 across three runs trained on three random samples of 500 prompts.

\section{Data Interventions}
\label{app:interventions}

Following \cite{kazemi2025big}, we design the given 7 interventions to improve data quality (Table~\ref{tab:interventions}) and prompt GPT-5-mini with given prompt. Since, we want to preserve the ground-truth answer, we apply these interventions only to the problem statement. Finally, to ensure the that the final answer is preserved, filter the augmented data with based on win-rate computed again with the augmented problem statements. In our experiments, we combine the original data and the intervened data in a ratio of 1:1. 
\input{tables/interventions}
\input{tables/intervention_results}

\section{Task Performance Analysis.}
\label{app:task_analysis}
\begin{figure}
  \centering
  \includegraphics[width=\linewidth]{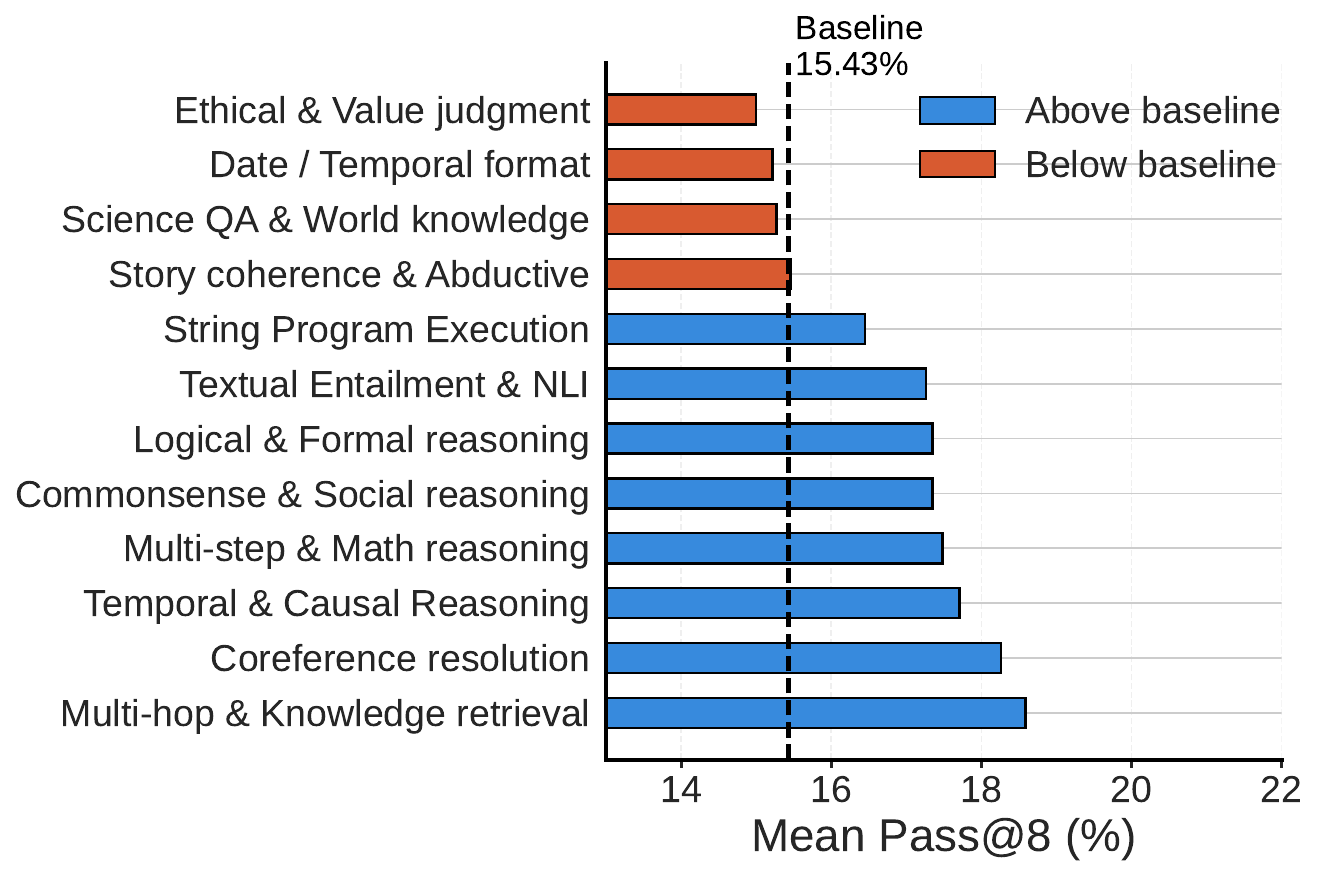}
    \caption{(a)We categorize the source tasks based on target reasoning skill and task type. We report Mean Pass@8 across each task category and highlight the categories which degrade baseline (Qwen3-0.6b) model on BBEH-mini.}
  \label{fig:task_analysis}
\end{figure}

We prompt an LLM (Claude-Opus-4.6) with the task descriptions from each task and generate coarse category labels. We find that Multi-hop Reasoning and Coreference resolution emerge as the strongest categories, while narrative and surface-formatting tasks  (e.g., Story Coherence, Date/Temporal format) consistently underperform (Figure~\ref{fig:task_analysis}). However, these aggregate trends obscure variations at the task-level. Despite Textual Entailment \& NLI ranking in the middle at the category-level, \texttt{task738\_perspectrum} 
emerges as the top-ranked task by large margin. This highlights that coarse category labels are insufficient for task selection and effective data curation for RL should be driven by fine-grained task utility analysis.

\section{Micro Mixing}
We provide the top tasks ranked per sub-task in Table~\ref{tab:top5_per_bbeh}. 
For Micro-Top1, 16 unique tasks are selected while 31 unique tasks are included in Micro-Top2. Additionally, we show the distribution of reasoning skills as categorized in \S~\ref{app:task_analysis} in \ours{} which is scaled from Micro-Top2 and comprises 31 unique tasks. 

\onecolumn
\begin{figure}
    \centering
    \includegraphics[width=\linewidth]{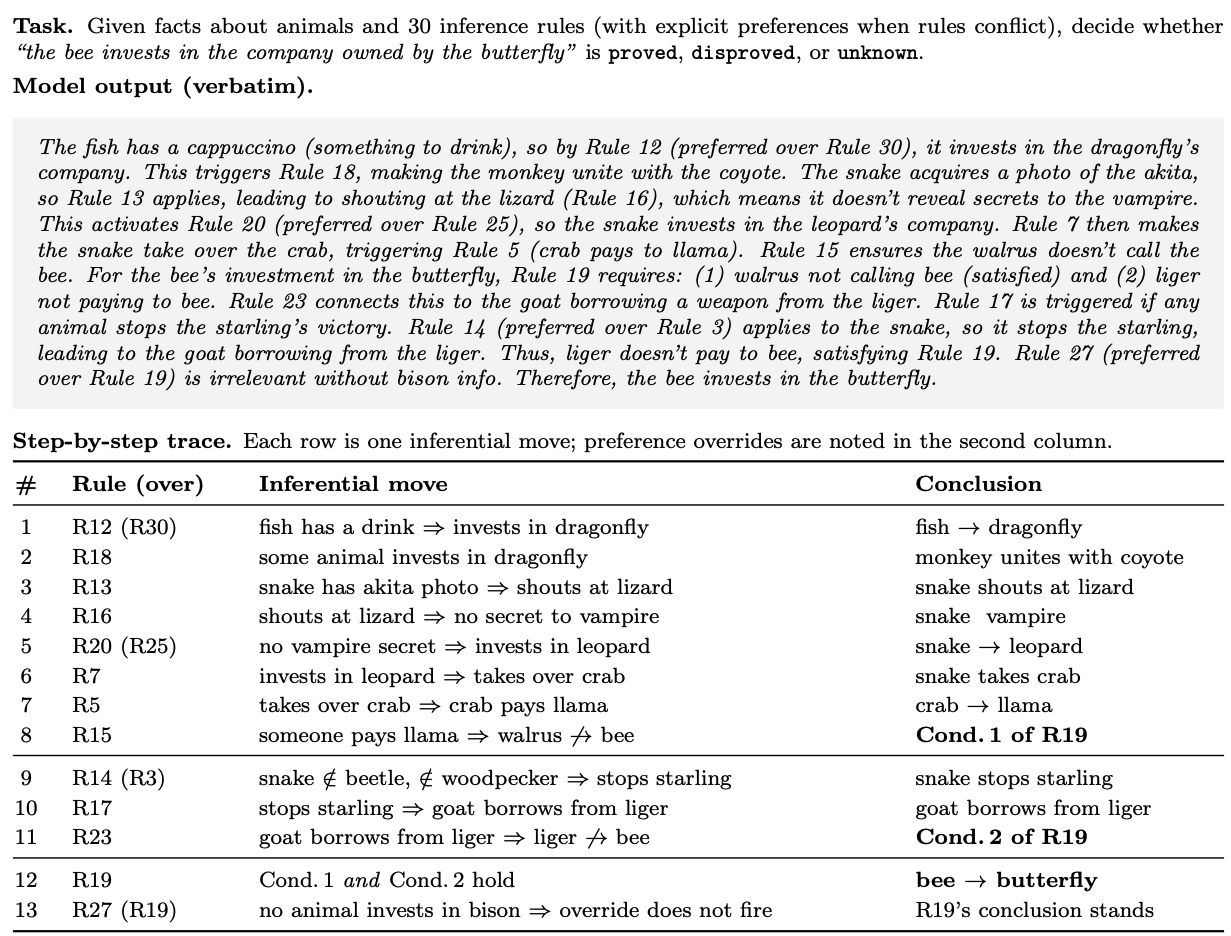}
    \caption{Verbatim output from \ours{}-4B on a Boardgame-QA instance with 30 rules and preference orderings,
together with the step-by-step inference trace it implies. The model produces the correct answer (proved)
by chaining 13 rule applications and correctly resolves four rule-preference conflicts.}
    \label{fig:cot}
\end{figure}

\begin{tcolorbox}[
    breakable,
    width=\textwidth,
    colback=white,
    colframe=headergreen,
    title={\centering \textbf{Prompt for Data Intervention}}
]
\footnotesize
\begin{lstlisting}[
    breaklines=true,
    postbreak={},
    breakindent=0pt,
    label={lst:intervention_prompt},
    frame=none,
    numbers=left,              % Enable line numbers
    numbersep=15pt,            % Space between numbers and code
    xleftmargin=20pt,          % Left margin for code alignment
    stepnumber=1               % Show every line number
]
You are an expert data augmentation assistant. Your task is to take an existing (problem, answer) pair from an NLP dataset and inject a distractor into the problem.
The distractor must make the problem harder for an AI model to solve, but it must NOT change the ground-truth answer.

## DISTRACTOR TYPE
You MUST use the following distractor type:
**{distractor_name}**: {distractor_description}

## RULES
1. **Answer preservation (CRITICAL)**: The ground-truth answer MUST remain exactly the same after distractor injection. Do not alter the core reasoning chain.
2. **Naturalness**: The distractor must read naturally within the problem. It should not feel artificially inserted or out of place.
3. **Plausibility**: The distractor should be plausible and contextually relevant enough that a model might be misled by it.
4. **Minimal invasion**: Modify only what is necessary. Do not rewrite the entire problem. Inject the distractor into or around the existing text.
5. **Difficulty calibration**: The distractor should make the problem meaningfully harder, not trivially so. Aim for a difficulty increase that would cause a mid-tier model to fail while a strong model would still succeed.

## OUTPUT FORMAT
You MUST respond with a valid JSON object and nothing else. No markdown, no explanation outside the JSON.
Use the following schema:
{{
  "original_problem": "",
  "original_solution": "",
  "augmented_problem": "",
  "augmented_solution": "",

  "distractor_metadata": {{
    "distractor_types_used": [
      {{
        "name": "",
        "description": ""
      }}
    ],
    "injected_text_summary": "",
    "why_answer_unchanged": "",
    "estimated_difficulty_increase": ""
  }}
}}

## IMPORTANT GUIDELINES
- Think step by step before generating the output.
- First, understand what the problem is asking and why the given answer is correct.
- Second, identify which parts of the problem can be augmented without breaking the answer.
- Third, apply the specified distractor type as naturally as possible.
- Fourth, draft the distractor text.
- Fifth, verify that the answer is still correct with the distractor in place.
- Only then produce the final JSON output.

## FINAL CHECKLIST (verify before outputting)
- [ ] Is the output valid JSON?
- [ ] Is the answer in augmented_problem identical to the original answer?
- [ ] Does the distractor read naturally in context?
- [ ] Is the specified distractor type used with a clear description?
- [ ] Is the why_answer_unchanged field filled with a logical explanation?
- [ ] Would the augmented problem genuinely be harder for a model to solve?

Now, process the following input and return the JSON output:

Problem: Statement: {problem}
Solution: {solution}
"""
\end{lstlisting}
\end{tcolorbox}

\onecolumn
\begin{tcolorbox}[
    breakable,
    width=\textwidth,
    colback=white,
    colframe=headergreen,
    title={\centering \textbf{Prompt for Reformatting Instruction-Tuning Dataset}}
]
\footnotesize
\begin{lstlisting}[
    breaklines=true,
    postbreak={},
    breakindent=0pt,
    label={lst:reformatting_prompt},
    frame=none,
    numbers=left,              % Enable line numbers
    numbersep=15pt,            % Space between numbers and code
    xleftmargin=20pt,          % Left margin for code alignment
    stepnumber=1               % Show every line number
]
Role: You are an expert Dataset Engineer specializing in Reinforcement Learning from Human Feedback (RLHF) and Verifiable Rewards.

Objective: Transform a raw task description, input, and output into a structured Problem and Solution pair. This pair must be suitable for RL training where the reward is calculated via exact-match verification.

Constraints:
The Problem: Must incorporate all necessary context from the input and output without giving away the output.
The Solution: Must contain only the final answer. No explanations, no "The answer is...", and no punctuation unless it is part of the value.
Verifiability: The solution must be uniquely extractable via simple string matching or regex.

Formatting Logic:
Open-ended: Use this if the answer is a unique value (e.g., a number, a specific name, or a constant).
MCQ (Multiple Choice): Use this if the task is subjective, has multiple valid answers, or involves Yes/No.
MCQ Format: Provide options labeled (A) through (J). If the task is multi-correct, the solution should be a comma-separated list of letters (e.g., "A, C").
Output Format: Return a valid JSON object with the keys "formatting logic", "problem" and "solution".

Task Description: {def_task}
Input: {example_input}
Output: {example_output}
\end{lstlisting}
\end{tcolorbox}
Finally, for win-rate filtering we generate 8 samples from Qwen3-0.6B at temperature=0.7 (generation length: 4096) , our baseline model and compute the per-question win-rate across these 8 samples. We filter all questions with a win-rate of 0 (too hard) and a win-rate of 1(too easy).

\input{tables/ranked_micro}

\input{tables/tasks}

%% file: sections/03-related_work.tex
\section{Related Work}

\subsection{General-Purpose Reasoning in LLMs}
Several works have explored expanding the general reasoning capabilities of LLMs. \cite{ma2025general} constructs a large-scale dataset spanning multiple domains such as history, finance, and physics from web-scraped sources. \cite{akter2026nemotron} goes beyond mathematics by curating synthetically derived questions from CommonCrawl and open-source QA datasets. \cite{lu2026goldengoosesimpletrick} leverages transformed pretraining data with structured templates and distractors to generate verifiable reasoning data in domains such as cybersecurity. However, these approaches largely rely on internet-sourced data, which can be noisy and of low quality. Other work has focused on rule-based tasks~\citep{liu2026rulereasonerreinforcedrulebasedreasoning} and logic puzzles~\citep{liu2025synlogicsynthesizingverifiablereasoning}. \citep{cheng2026revisiting} employs data mixing across diverse domains such as math, logic and tabular and studies data curation for RLVR, however, they rely on domain-specific rewards and complex data filtering to combine data from existing domain specific datasets. While effective for specialized reasoning, these approaches rely on highly specialized domain-specific datasets that are challenging to scale and require complex reward design and filtering.  In contrast, \ours{} leverages instruction-tuning datasets, which are human-annotated and can be made usable for RLVR with simple reformatting and scale easily across diverse reasoning types. 

\subsection{Data Curation for Reasoning}
High-quality reasoning data is critical for training strong LLM reasoners. Prior work has focused on large-scale datasets for supervised fine-tuning (SFT) \citep{openr1,zhao20251} and RLVR \citep{chen2025acereason, hu2025open}, typically by scraping competition websites or distilling knowledge from larger models. On the other hand, \citep{muennighoff2025s1, ye2025limo} demonstrate that carefully curated, high-quality reasoning datasets can yield strong gains even with relatively small datasets. \cite{guha2025openthoughts} systematically studies data design principles for SFT reasoning data at scale through controlled experiments, in a manner similar to \ours{}. However, these efforts primarily focus on reasoning in formal domains using SFT. SFT aims to improve instruction-following by training on demonstrations~\citep{zhang2025instructiontuninglargelanguage, wang2022super}, while RLVR optimizes a sparse, outcome-based reward~\citep{guo2025deepseek}. Moreover, SFT typically requires complete reasoning traces and solutions for training, whereas RLVR only requires the final answer. As a result, SFT-oriented data curation strategies do not directly transfer to RLVR. \ours{} addresses this gap by providing key insights to drive data curation for RLVR.

%% file: tables/supernova_egs.tex
\begin{table*}[t]
\centering
\small
\renewcommand{\arraystretch}{1.2}
\begin{tabularx}{\textwidth}{@{}p{0.13\textwidth} X@{}}
\toprule
\multicolumn{2}{@{}l}{\textbf{Task A: \texttt{task738\_perspectrum\_classification} --- claim/perspective stance detection (classification, support/undermine).}} \\
\midrule
\multicolumn{2}{@{}l}{\textit{Shared raw task definition:} ``In this task you will be given a claim and a perspective. You should determine whether that}\\
\multicolumn{2}{@{}l}{perspective supports or undermines the claim. If the perspective could possibly convince someone with different view,}\\
\multicolumn{2}{@{}l}{it is supporting, otherwise it is undermining.''} \\
\midrule

\rowcolor{gray!12}
\textbf{Raw} & claim: Children should not be allowed to inherit vast wealth as this damages them and society. \newline perspective: Inherited wealth demotivates the recipients so that they put less effort into training, education and social skills. \newline\textbf{Output:} \texttt{support} \\
\textbf{Reformat} \newline \textit{(MCQ, 2-way)} & Claim: Children should not be allowed to inherit vast wealth as this damages them and society. Perspective: Inherited wealth demotivates the recipients so that they put less effort into training, education and social skills. Question: Does the perspective support or undermine the claim? Choose one option: (A) Supports \quad (B) Undermines \newline\textbf{Answer:} \texttt{A} \\
\midrule

\rowcolor{gray!12}
\textbf{Raw} & claim: Domestic intelligence agencies have a legitimate role to play in democracy. \newline perspective: The government does not have the right to spy on its citizens. \newline\textbf{Output:} \texttt{undermine} \\
\textbf{Reformat} \newline \textit{(open-ended)} & Claim: Domestic intelligence agencies have a legitimate role to play in democracy. Perspective: The government does not have the right to spy on its citizens. Rule: If the perspective could possibly convince someone with a different view, it is supporting; otherwise it is undermining. Question: Does the perspective support or undermine the claim? \textbf{Answer with a single word: support or undermine.} \newline\textbf{Answer:} \texttt{undermine} \\
\midrule

\rowcolor{gray!12}
\textbf{Raw} & claim: Marriage is an outdated institution. \newline perspective: Those who are observant religiously think that marriage is important. \newline\textbf{Output:} \texttt{undermine} \\
\textbf{Reformat} \newline \textit{(MCQ, 10-way)} & Determine whether the perspective supports or undermines the claim. Claim: Marriage is an outdated institution. Perspective: Those who are observant religiously think that marriage is important. Choose the best option: (A) Supports (B) Undermines (C) Both supports and undermines (D) Neither (E) Neutral or irrelevant (F) Depends on context (G) Ambiguous (H) Contradicts (I) Unrelated (J) Cannot determine \newline\textbf{Answer:} \texttt{B} \\

\bottomrule
\end{tabularx}
\caption{\textbf{Examples from \ours{} dataset.} We show examples from our top performing task task738. In the reformatting step of our pipeline, we employs LLMs to (i) inline the shared task description in each sample, (ii) produce both open-ended and multiple-choice variants, and (iii) vary MCQ option counts from 2-way to 10-way. This reformatting across diverse tasks, normalizes the answer format and allows for rule-base verification.}
\label{tab:reformat-example}
\end{table*}

%% file: tables/interventions.tex
\begin{table*}[ht]
\centering

\begin{tabular}{p{4cm} p{10cm}}
\toprule
\textbf{Dimension} & \textbf{Description} \\
\midrule
Many-hop reasoning & Add information that increases the number of reasoning steps needed to reach the answer. \\
Going against strong prior & Add context that creates a misleading prior belief which conflicts with the correct answer, tempting the model to answer incorrectly based on surface-level associations. \\
Learning on the fly & Introduce a new rule, definition, or convention within the problem that must be understood and applied to solve it. \\
Long-context & Pad the problem with additional (but non-answer-changing) context to increase overall length. \\
Finding errors in reasoning traces & Include a flawed reasoning chain within the problem that the model must recognize as incorrect. \\
Inductive reasoning & Provide a set of examples that establish a pattern, requiring the model to induce and apply the pattern. \\
Constraint satisfaction & Add extra constraints that the model must track, even though they do not affect the final answer. \\
Compositional understanding & Fuse an independent sub-problem into the main problem, requiring the model to separate and solve them independently. \\
Knowledge-intensive reasoning & Add domain-specific terminology or context that requires specialized knowledge to parse, even though it does not change the answer. \\
\bottomrule
\end{tabular}
\caption{Following \cite{kazemi2025big}, we design these interventions to improve the data quality. We provide the interventions and their definitions here.}
\label{tab:interventions}
\end{table*}

%% file: tables/intervention_results.tex
\begin{table}[]
    \centering
    \begin{tabular}{lc}
    \toprule
    \textbf{Intervention} & \textbf{pass@8} \\
    \midrule
    \rowcolor{blue!8} Micro-Top2 & \textbf{22.8} \\
    Going Against Prior & 22.6 \\
    Long-Context & 21.3 \\
    Inductive Reasoning & 20.4 \\
    Finding Errors & 20.0 \\
    Many-hop Reasoning & 20.0 \\
    Knowledge-intensive Reasoning & 19.8 \\
    Compositional Understanding & 19.6 \\
    Learning on the Fly & 18.3 \\
    \bottomrule
  
  \end{tabular}
     \caption{\textbf{Impact of interventions.} We compare the
    performance of models trained on datasets transformed using
    synthetic interventions. We find that the base dataset
    is superior to all interventions.}
     \label{tab:interventions_results}
\end{table}

%% file: tables/ranked_micro.tex
{\small
\setlength{\tabcolsep}{3pt}
\begin{longtable}{@{} p{3.4cm} c p{1.6cm} p{\dimexpr\textwidth-3.2cm-1.2cm-1.6cm-12pt-6\tabcolsep\relax} @{}}
\caption{Top-5 training tasks per BBEH task.}
\label{tab:top5_per_bbeh} \\
\toprule
\textbf{BBEH Task} & \textbf{Rank} & \textbf{Task ID} & \textbf{Training Task} \\
\midrule
\endfirsthead

\multicolumn{4}{c}{\tablename\ \thetable{} -- \textit{continued from previous page}} \\[4pt]
\toprule
\textbf{BBEH Task} & \textbf{Rank} & \textbf{Task ID} & \textbf{Training Task} \\
\midrule
\endhead

\midrule
\multicolumn{4}{r}{\textit{continued on next page}} \\
\endfoot

\bottomrule
\endlastfoot
movie recommendation & 1 & \texttt{task827} & copa\_commonsense\_reasoning \\
 & 2 & \texttt{task069} & abductivenli\_classification \\
 & 3 & \texttt{task212} & logic2text\_classification \\
 & 4 & \texttt{task1297} & qasc\_question\_answering \\
 & 5 & \texttt{task1209} & atomic\_classification\_objectuse \\
\midrule
word sorting & 1 & \texttt{task828} & copa\_commonsense\_cause\_effect \\
 & 2 & \texttt{task1548} & wiqa\_binary\_classification \\
 & 3 & \texttt{task1385} & anli\_r1\_entailment \\
 & 4 & \texttt{task835} & mathdataset\_answer\_generation \\
 & 5 & \texttt{task383} & matres\_classification \\
\midrule
object counting & 1 & \texttt{task1210} & atomic\_classification\_madeupof \\
 & 2 & \texttt{task1211} & atomic\_classification\_hassubevent \\
 & 3 & \texttt{task1155} & bard\_analogical\_reasoning\_trash\_or\_treasure \\
 & 4 & \texttt{task827} & copa\_commonsense\_reasoning \\
 & 5 & \texttt{task004} & mctaco\_answer\_generation\_event\_duration \\
\midrule
geometric shapes & 1 & \texttt{task249} & enhanced\_wsc\_pronoun\_disambiguation \\
 & 2 & \texttt{task1209} & atomic\_classification\_objectuse \\
 & 3 & \texttt{task1385} & anli\_r1\_entailment \\
 & 4 & \texttt{task697} & mmmlu\_answer\_generation\_formal\_logic \\
 & 5 & \texttt{task717} & mmmlu\_answer\_generation\_logical\_fallacies \\
\midrule
nycc & 1 & \texttt{task1297} & qasc\_question\_answering \\
 & 2 & \texttt{task073} & commonsenseqa\_answer\_generation \\
 & 3 & \texttt{task212} & logic2text\_classification \\
 & 4 & \texttt{task213} & rocstories\_correct\_ending\_classification \\
 & 5 & \texttt{task828} & copa\_commonsense\_cause\_effect \\
\midrule
boardgame qa & 1 & \texttt{task004} & mctaco\_answer\_generation\_event\_duration \\
 & 2 & \texttt{task116} & com2sense\_commonsense\_reasoning \\
 & 3 & \texttt{task062} & bigbench\_repeat\_copy\_logic \\
 & 4 & \texttt{task1726} & mathqa\_correct\_answer\_generation \\
 & 5 & \texttt{task1387} & anli\_r3\_entailment \\
\midrule
buggy tables & 1 & \texttt{task007} & mctaco\_answer\_generation\_transient\_stationary \\
 & 2 & \texttt{task1390} & wscfixed\_coreference \\
 & 3 & \texttt{task600} & find\_the\_longest\_common\_substring\_in\_two\_strings \\
 & 4 & \texttt{task391} & causal\_relationship \\
 & 5 & \texttt{task004} & mctaco\_answer\_generation\_event\_duration \\
\midrule
linguini & 1 & \texttt{task004} & mctaco\_answer\_generation\_event\_duration \\
 & 2 & \texttt{task1209} & atomic\_classification\_objectuse \\
 & 3 & \texttt{task640} & esnli\_classification \\
 & 4 & \texttt{task085} & unnatural\_addsub\_arithmetic \\
 & 5 & \texttt{task738} & perspectrum\_classification \\
\midrule
boolean expressions & 1 & \texttt{task850} & synthetic\_longest\_palindrome \\
 & 2 & \texttt{task600} & find\_the\_longest\_common\_substring\_in\_two\_strings \\
 & 3 & \texttt{task1390} & wscfixed\_coreference \\
 & 4 & \texttt{task018} & mctaco\_temporal\_reasoning\_presence \\
 & 5 & \texttt{task210} & logic2text\_structured\_text\_generation \\
\midrule
multistep arithmetic & 1 & \texttt{task004} & mctaco\_answer\_generation\_event\_duration \\
 & 2 & \texttt{task1210} & atomic\_classification\_madeupof \\
 & 3 & \texttt{task1211} & atomic\_classification\_hassubevent \\
 & 4 & \texttt{task007} & mctaco\_answer\_generation\_transient\_stationary \\
 & 5 & \texttt{task1390} & wscfixed\_coreference \\
\midrule
time arithmetic & 1 & \texttt{task212} & logic2text\_classification \\
 & 2 & \texttt{task835} & mathdataset\_answer\_generation \\
 & 3 & \texttt{task383} & matres\_classification \\
 & 4 & \texttt{task1209} & atomic\_classification\_objectuse \\
 & 5 & \texttt{task1153} & bard\_analogical\_reasoning\_affordance \\
\midrule
object properties & 1 & \texttt{task828} & copa\_commonsense\_cause\_effect \\
 & 2 & \texttt{task004} & mctaco\_answer\_generation\_event\_duration \\
 & 3 & \texttt{task1210} & atomic\_classification\_madeupof \\
 & 4 & \texttt{task1211} & atomic\_classification\_hassubevent \\
 & 5 & \texttt{task007} & mctaco\_answer\_generation\_transient\_stationary \\
\midrule
hyperbaton & 1 & \texttt{task249} & enhanced\_wsc\_pronoun\_disambiguation \\
 & 2 & \texttt{task213} & rocstories\_correct\_ending\_classification \\
 & 3 & \texttt{task393} & plausible\_result\_generation \\
 & 4 & \texttt{task600} & find\_the\_longest\_common\_substring\_in\_two\_strings \\
 & 5 & \texttt{task827} & copa\_commonsense\_reasoning \\
\midrule
sarc triples & 1 & \texttt{task210} & logic2text\_structured\_text\_generation \\
 & 2 & \texttt{task640} & esnli\_classification \\
 & 3 & \texttt{task970} & sherliic\_causal\_relationship \\
 & 4 & \texttt{task850} & synthetic\_longest\_palindrome \\
 & 5 & \texttt{task717} & mmmlu\_answer\_generation\_logical\_fallacies \\
\midrule
zebra puzzles & 1 & \texttt{task828} & copa\_commonsense\_cause\_effect \\
 & 2 & \texttt{task863} & asdiv\_multiop\_question\_answering \\
 & 3 & \texttt{task210} & logic2text\_structured\_text\_generation \\
 & 4 & \texttt{task1726} & mathqa\_correct\_answer\_generation \\
 & 5 & \texttt{task738} & perspectrum\_classification \\
\midrule
spatial reasoning & 1 & \texttt{task738} & perspectrum\_classification \\
 & 2 & \texttt{task087} & new\_operator\_addsub\_arithmetic \\
 & 3 & \texttt{task019} & mctaco\_temporal\_reasoning\_category \\
 & 4 & \texttt{task080} & piqa\_answer\_generation \\
 & 5 & \texttt{task697} & mmmlu\_answer\_generation\_formal\_logic \\
\midrule
shuffled objects & 1 & \texttt{task249} & enhanced\_wsc\_pronoun\_disambiguation \\
 & 2 & \texttt{task018} & mctaco\_temporal\_reasoning\_presence \\
 & 3 & \texttt{task827} & copa\_commonsense\_reasoning \\
 & 4 & \texttt{task1297} & qasc\_question\_answering \\
 & 5 & \texttt{task717} & mmmlu\_answer\_generation\_logical\_fallacies \\
\midrule
temporal sequence & 1 & \texttt{task004} & mctaco\_answer\_generation\_event\_duration \\
 & 2 & \texttt{task1210} & atomic\_classification\_madeupof \\
 & 3 & \texttt{task1211} & atomic\_classification\_hassubevent \\
 & 4 & \texttt{task007} & mctaco\_answer\_generation\_transient\_stationary \\
 & 5 & \texttt{task1390} & wscfixed\_coreference \\
\midrule
sportqa & 1 & \texttt{task270} & csrg\_counterfactual\_context\_generation \\
 & 2 & \texttt{task210} & logic2text\_structured\_text\_generation \\
 & 3 & \texttt{task062} & bigbench\_repeat\_copy\_logic \\
 & 4 & \texttt{task600} & find\_the\_longest\_common\_substring\_in\_two\_strings \\
 & 5 & \texttt{task391} & causal\_relationship \\
\midrule
web of lies & 1 & \texttt{task1152} & bard\_analogical\_reasoning\_causation \\
 & 2 & \texttt{task828} & copa\_commonsense\_cause\_effect \\
 & 3 & \texttt{task1211} & atomic\_classification\_hassubevent \\
 & 4 & \texttt{task080} & piqa\_answer\_generation \\
 & 5 & \texttt{task640} & esnli\_classification \\
\midrule
causal understanding & 1 & \texttt{task383} & matres\_classification \\
 & 2 & \texttt{task004} & mctaco\_answer\_generation\_event\_duration \\
 & 3 & \texttt{task1390} & wscfixed\_coreference \\
 & 4 & \texttt{task828} & copa\_commonsense\_cause\_effect \\
 & 5 & \texttt{task291} & semeval\_2020\_task4\_commonsense\_validation \\
\midrule
disambiguation qa & 1 & \texttt{task697} & mmmlu\_answer\_generation\_formal\_logic \\
 & 2 & \texttt{task1296} & wiki\_hop\_question\_answering \\
 & 3 & \texttt{task717} & mmmlu\_answer\_generation\_logical\_fallacies \\
 & 4 & \texttt{task018} & mctaco\_temporal\_reasoning\_presence \\
 & 5 & \texttt{task065} & timetravel\_consistent\_sentence\_classification \\
\midrule
dyck languages & 1 & \texttt{task1390} & wscfixed\_coreference \\
 & 2 & \texttt{task1386} & anli\_r2\_entailment \\
 & 3 & \texttt{task004} & mctaco\_answer\_generation\_event\_duration \\
 & 4 & \texttt{task1210} & atomic\_classification\_madeupof \\
 & 5 & \texttt{task1152} & bard\_analogical\_reasoning\_causation \\
\end{longtable}
}

%% file: tables/tasks.tex
\begin{longtable}{@{}>{\raggedright\arraybackslash}p{0.38\textwidth}>{\raggedright\arraybackslash}p{0.58\textwidth}@{}}
\caption{Task descriptions.}
\label{tab:tasks} \\  
\toprule
\textbf{Task Name} & \textbf{Summary} \\  
\midrule
\endfirsthead
\toprule
\textbf{Task Name} & \textbf{Summary} \\  
\midrule
\endhead
\midrule
\multicolumn{2}{r}{\textit{Continued on next page}} \\ 
\endfoot
\bottomrule
\endlastfoot

task738 perspectrum classification & Decide whether the given perspective supports or undermines the given claim. \\ 
\hline
task003 mctaco question generation event duration & Writing questions that involve commonsense understanding of ``event duration''. \\  \hline
task717 mmmlu answer generation logical fallacies & Answering multiple choice questions on logical fallacies. \\\hline
task249 enhanced wsc pronoun disambiguation & Given a sentence and a pronoun, decide which one of the choices the pronoun is referring to. \\  \hline
task1385 anli r1 entailment & Given a premise and hypothesis, determine if the hypothesis entails, contradicts, or is neutral to the premise. \\  \hline
task1296 wiki hop question answering & Given a subject, a relation, and a context, find the object with that relation to the subject. \\  \hline
task828 copa commonsense cause effect & Given a pair of sentences, judge whether the second sentence is the cause or effect of the first one. \\  \hline
task073 commonsenseqa answer generation & Answer questions based on commonsense knowledge. \\  \hline
task018 mctaco temporal reasoning presence & Checking the presence of temporal reasoning in a question. \\  \hline
task697 mmmlu answer generation formal logic & Answering multiple choice questions on formal logic. \\  \hline
task827 copa commonsense reasoning & Given a premise and two alternatives, select the alternative that more plausibly has a causal relation with the premise. \\  \hline
task383 matres classification & Given a context and a verb, answer if the given verb can be anchored in time or not. \\  \hline
task065 timetravel consistent sentence classification & Choosing the option that makes a given short story consistent. \\  \hline
task640 esnli classification & Given a premise and hypothesis, determine if the hypothesis entails, contradicts, or is neutral to the premise. \\  \hline
task1387 anli r3 entailment & Given a premise and hypothesis, determine if the hypothesis entails, contradicts, or is neutral to the premise. \\  \hline
task863 asdiv multiop question answering & Given a mathematical question involving multiple operations, find the most suitable numerical answer. \\  \hline
task1209 atomic classification objectuse & Given a tuple, determine whether the Head is used for the Tail or not. \\  \hline
task212 logic2text classification & Given a command, classify the command in one of seven logic types. \\  \hline
task750 aqua multiple choice answering & Given a mathematical question, find the most suitable numerical answer. \\  \hline
task010 mctaco answer generation event ordering & Answering questions that involve commonsense understanding of event ordering. \\  \hline
task1297 qasc question answering & Given two facts and a multiple-choice question, answer the question. \\  \hline
task007 mctaco answer generation transient stationary & Answering questions that involve commonsense understanding of transient vs.\ stationary events. \\  \hline
task1390 wscfixed coreference & Given a context, a pronoun, and a noun, determine if the pronoun in the context refers to the noun or not. \\  \hline
task600 find the longest common substring in two strings & Given two strings return the longest common substring in those two strings. \\  \hline
task080 piqa answer generation & Generate a solution to a goal regarding physical knowledge about the world. \\  \hline
task1726 mathqa correct answer generation & Generate correct answers for math questions. \\  \hline
task835 mathdataset answer generation & Find the numerical answer for a math word problem. \\  \hline
task580 socialiqa answer generation & Given a context, a question and three options, provide the correct answer based on the context. \\  \hline
task1393 superglue copa text completion & Given a premise sentence, two possible options and a question word, choose the best option. \\  \hline
task1727 wiqa what is the effect & Find the effect of an event on another event, based on an introduced process. \\  \hline
task170 hotpotqa answer generation & Given a set of context and supporting facts, answer the question asked. \\  \hline
task133 winowhy reason plausibility detection & Detect if a reason that explains an answer to a pronoun coreference resolution question is correct or not. \\  \hline
task004 mctaco answer generation event duration & Answering questions that involve commonsense understanding of event duration. \\  \hline
task019 mctaco temporal reasoning category & Verifying the temporal reasoning category of a given question. \\  \hline
task229 arc answer generation hard & Given a hard science question, provide the answer based on scientific facts and reasoning. \\  \hline
task106 scruples ethical judgment & Given two actions choose the one that is considered less ethical. \\  \hline
task178 quartz question answering & Given a question, select the correct answer from the given options using an explanation. \\  \hline
task1152 bard analogical reasoning causation & Given an analogy that relates actions with their consequences, give the appropriate consequence of the given action. \\  \hline
task090 equation learner algebra & Answer the given equation. \\  \hline
task850 synthetic longest palindrome & Given a string find the longest substring that is a palindrome. \\  \hline
task1422 mathqa physics & Given a problem on physics and options to choose from, find the correct option that answers the problem. \\  \hline
task393 plausible result generation & Given a sentence, write another sentence that is a likely result of it. \\  \hline
task085 unnatural addsub arithmetic & Performing arithmetic with swapped operator symbols. \\  \hline
task1529 scitail1.1 classification & Determining if there is entailment between hypothesis and premise. \\  \hline
task867 mawps multiop question answering & Given a mathematical question involving multiple operations, find the most suitable numerical answer. \\  \hline
task211 logic2text classification & Given a command and corresponding interpretation, classify whether it is the right interpretation or not. \\  \hline
task1548 wiqa binary classification & Binary classification based on steps in wiqa. \\  \hline
task966 ruletaker fact checking based on given context & Fact checking based on given context. \\  \hline
task935 defeasible nli atomic classification & Given a premise, hypothesis and an update, identify whether the update strengthens or weakens the hypothesis. \\  \hline
task116 com2sense commonsense reasoning & Decide whether a sentence is plausible and matches commonsense. \\  \hline
task087 new operator addsub arithmetic & Performing arithmetic with newly defined operator symbols. \\  \hline
task206 collatz conjecture & Given a list of integers, compute the next number in the 3n+1 problem. \\  \hline
task970 sherliic causal relationship & Determine if A and B share a causal relationship. \\  \hline
task086 translated symbol arithmetic & Performing arithmetic with translated operator symbols. \\  \hline
task270 csrg counterfactual context generation & Given a premise, initial context with ending, and new counterfactual ending, generate counterfactual context which supports the new story ending. \\  \hline
task392 inverse causal relationship & Given two sentences, decide whether the first sentence can be the result of the second one. \\  \hline
task105 story cloze-rocstories sentence generation & Given four sentences, predict the next coherent sentence. \\  \hline
task1507 boolean temporal reasoning & Given a statement about date and time values, deduce whether it is true or false. \\  \hline
task1404 date conversion & Given a date in a particular format, convert it into some other format. \\  \hline
task1153 bard analogical reasoning affordance & Given an analogy that signifies affordances, give the appropriate affordance of the given action. \\  \hline
task069 abductivenli classification & Choosing text that completes a story based on given beginning and ending. \\  \hline
task062 bigbench repeat copy logic & Generating text that follows simple logical operations such as repeat, before, after etc. \\  \hline
task1088 array of products & Given an integer array, return an array such that its element at each location is equal to the product of elements at every other location in the input array. \\  \hline
task190 snli classification & Given two sentences choose whether they agree, disagree, or neither with each other. \\  \hline
task1333 check validity date ddmmyyyy & Given a date in dd/mm/yyyy format, check if it is a valid date or not. \\  \hline
task016 mctaco answer generation frequency & Answering questions that involve commonsense understanding of event frequency. \\  \hline
task1208 atomic classification xreason & Given a tuple, determine whether the Tail is the reason for the Head or not. \\  \hline
task1386 anli r2 entailment & Given a premise and hypothesis, determine if the hypothesis entails, contradicts, or is neutral to the premise. \\  \hline
task1516 imppres naturallanguageinference & Classify a given premise and hypothesis pair. \\  \hline
task199 mnli classification & Given 2 sentences, determine if they clearly agree or disagree with each other or if they cannot be answered. \\  \hline
task1210 atomic classification madeupof & Given a tuple, determine whether the Head is made of the Tail or not. \\  \hline
task217 rocstories ordering answer generation & Given a five sentence story in shuffled order and the title, put the story in the correct order. \\  \hline
task1155 bard analogical reasoning trash or treasure & Given an analogy that relates items to whether they are trash or treasure, determine if the given item is trash or treasure. \\  \hline
task218 rocstories swap order answer generation & Given a five sentence story and the title, determine which two sentences must be swapped so that the story makes complete sense. \\  \hline
task1211 atomic classification hassubevent & Given a tuple, determine whether the Head includes an event or an action in the Tail or not. \\  \hline
task213 rocstories correct ending classification & Given the title and the first four sentences of a five sentence story, choose the correct story ending. \\  \hline
\label{app_tab:superni}
\end{longtable}

%% file: custom.bib
@article{bansal2025honeybee,
  title={Honeybee: Data recipes for vision-language reasoners},
  author={Bansal, Hritik and Sachan, Devandra Singh and Chang, Kai-Wei and Grover, Aditya and Ghosh, Gargi and Yih, Wen-tau and Pasunuru, Ramakanth},
  journal={arXiv preprint arXiv:2510.12225},
  year={2025}
}

@article{wei2021finetuned,
  title={Finetuned language models are zero-shot learners},
  author={Wei, Jason and Bosma, Maarten and Zhao, Vincent Y and Guu, Kelvin and Yu, Adams Wei and Lester, Brian and Du, Nan and Dai, Andrew M and Le, Quoc V},
  journal={arXiv preprint arXiv:2109.01652},
  year={2021}
}

@misc{hendrycks2021measuringmathematicalproblemsolving,
      title={Measuring Mathematical Problem Solving With the MATH Dataset}, 
      author={Dan Hendrycks and Collin Burns and Saurav Kadavath and Akul Arora and Steven Basart and Eric Tang and Dawn Song and Jacob Steinhardt},
      year={2021},
      eprint={2103.03874},
      archivePrefix={arXiv},
      primaryClass={cs.LG},
      url={https://arxiv.org/abs/2103.03874}, 
}

@article{yue2025limit-of-rlvr,
  title={Does Reinforcement Learning Really Incentivize Reasoning Capacity in LLMs Beyond the Base Model?},
  author={Yue, Yang and Chen, Zhiqi and Lu, Rui and Zhao, Andrew and Wang, Zhaokai and Yue, Yang and Song, Shiji and Huang, Gao},
  journal={arXiv preprint arXiv:2504.13837},
  year={2025}
}

@misc{aime24,
  title={American Invitational Mathematics Examination (AIME) 2024},
  author={Zhang, Yifan and Math-AI, Team},
  year={2024},
  howpublished={\url{https://huggingface.co/datasets/math-ai/aime24}}
}

@article{ahmad2025opencodereasoning,
      title={{OpenCodeReasoning: Advancing Data Distillation for Competitive Coding}}, 
      author={Ahmad, Wasi Uddin and Narenthiran, Sean and Majumdar, Somshubra and Ficek, Aleksander and Jain, Siddhartha and Huang, Jocelyn and Noroozi, Vahid and Ginsburg, Boris},
      year={2025},
      eprint={2504.01943},
      archivePrefix={arXiv},
      primaryClass={cs.CL},
      journal={arXiv preprint arXiv:2504.01943},
      url={https://arxiv.org/abs/2504.01943}, 
}

@article{olmo2025olmo,
  title={Olmo 3},
  author={Olmo, Team and Ettinger, Allyson and Bertsch, Amanda and Kuehl, Bailey and Graham, David and Heineman, David and Groeneveld, Dirk and Brahman, Faeze and Timbers, Finbarr and Ivison, Hamish and others},
  journal={arXiv preprint arXiv:2512.13961},
  year={2025}
}

@article{ye2025limo,
  title={Limo: Less is more for reasoning},
  author={Ye, Yixin and Huang, Zhen and Xiao, Yang and Chern, Ethan and Xia, Shijie and Liu, Pengfei},
  journal={arXiv preprint arXiv:2502.03387},
  year={2025}
}

@inproceedings{lightman2023let,
  title={Let's verify step by step},
  author={Lightman, Hunter and Kosaraju, Vineet and Burda, Yuri and Edwards, Harrison and Baker, Bowen and Lee, Teddy and Leike, Jan and Schulman, John and Sutskever, Ilya and Cobbe, Karl},
  booktitle={The twelfth international conference on learning representations},
  year={2023}
}

@article{chen2021evaluating,
  title={Evaluating large language models trained on code},
  author={Chen, Mark and Tworek, Jerry and Jun, Heewoo and Yuan, Qiming and Pinto, Henrique Ponde De Oliveira and Kaplan, Jared and Edwards, Harri and Burda, Yuri and Joseph, Nicholas and Brockman, Greg and others},
  journal={arXiv preprint arXiv:2107.03374},
  year={2021}
}

@article{grattafiori2024llama,
  title={The llama 3 herd of models},
  author={Grattafiori, Aaron and Dubey, Abhimanyu and Jauhri, Abhinav and Pandey, Abhinav and Kadian, Abhishek and Al-Dahle, Ahmad and Letman, Aiesha and Mathur, Akhil and Schelten, Alan and Vaughan, Alex and others},
  journal={arXiv preprint arXiv:2407.21783},
  year={2024}
}

@article{shao2024deepseekmath,
  title={Deepseekmath: Pushing the limits of mathematical reasoning in open language models},
  author={Shao, Zhihong and Wang, Peiyi and Zhu, Qihao and Xu, Runxin and Song, Junxiao and Bi, Xiao and Zhang, Haowei and Zhang, Mingchuan and Li, YK and Wu, Yang and others},
  journal={arXiv preprint arXiv:2402.03300},
  year={2024}
}

@inproceedings{muennighoff2025s1,
  title={s1: Simple test-time scaling},
  author={Muennighoff, Niklas and Yang, Zitong and Shi, Weijia and Li, Xiang Lisa and Fei-Fei, Li and Hajishirzi, Hannaneh and Zettlemoyer, Luke and Liang, Percy and Cand{\`e}s, Emmanuel and Hashimoto, Tatsunori B},
  booktitle={Proceedings of the 2025 Conference on Empirical Methods in Natural Language Processing},
  pages={20286--20332},
  year={2025}
}

@inproceedings{xiong2024large,
  title={Large language models can learn temporal reasoning},
  author={Xiong, Siheng and Payani, Ali and Kompella, Ramana and Fekri, Faramarz},
  booktitle={Proceedings of the 62nd Annual Meeting of the Association for Computational Linguistics (Volume 1: Long Papers)},
  pages={10452--10470},
  year={2024}
}

@article{team2026qwen3,
  title={Qwen3. 5: Towards native multimodal agents},
  author={Team, Qwen},
  journal={URL: https://qwen. ai/blog},
  year={2026}
}

@misc{liu2025synlogicsynthesizingverifiablereasoning,
      title={SynLogic: Synthesizing Verifiable Reasoning Data at Scale for Learning Logical Reasoning and Beyond}, 
      author={Junteng Liu and Yuanxiang Fan and Zhuo Jiang and Han Ding and Yongyi Hu and Chi Zhang and Yiqi Shi and Shitong Weng and Aili Chen and Shiqi Chen and Yunan Huang and Mozhi Zhang and Pengyu Zhao and Junjie Yan and Junxian He},
      year={2025},
      eprint={2505.19641},
      archivePrefix={arXiv},
      primaryClass={cs.AI},
      url={https://arxiv.org/abs/2505.19641}, 
}

@misc{zhang2025instructiontuninglargelanguage,
      title={Instruction Tuning for Large Language Models: A Survey}, 
      author={Shengyu Zhang and Linfeng Dong and Xiaoya Li and Sen Zhang and Xiaofei Sun and Shuhe Wang and Jiwei Li and Runyi Hu and Tianwei Zhang and Fei Wu and Guoyin Wang},
      year={2025},
      eprint={2308.10792},
      archivePrefix={arXiv},
      primaryClass={cs.CL},
      url={https://arxiv.org/abs/2308.10792}, 
}

@misc{lu2026goldengoosesimpletrick,
      title={Golden Goose: A Simple Trick to Synthesize Unlimited RLVR Tasks from Unverifiable Internet Text}, 
      author={Ximing Lu and David Acuna and Jaehun Jung and Jian Hu and Di Zhang and Shizhe Diao and Yunheng Zou and Shaokun Zhang and Brandon Cui and Mingjie Liu and Hyunwoo Kim and Prithviraj Ammanabrolu and Jan Kautz and Yi Dong and Yejin Choi},
      year={2026},
      eprint={2601.22975},
      archivePrefix={arXiv},
      primaryClass={cs.AI},
      url={https://arxiv.org/abs/2601.22975}, 
}

@misc{liu2026rulereasonerreinforcedrulebasedreasoning,
      title={RuleReasoner: Reinforced Rule-based Reasoning via Domain-aware Dynamic Sampling}, 
      author={Yang Liu and Jiaqi Li and Zilong Zheng},
      year={2026},
      eprint={2506.08672},
      archivePrefix={arXiv},
      primaryClass={cs.CL},
      url={https://arxiv.org/abs/2506.08672}, 
}

@article{yang2025qwen3,
  title={Qwen3 technical report},
  author={Yang, An and Li, Anfeng and Yang, Baosong and Zhang, Beichen and Hui, Binyuan and Zheng, Bo and Yu, Bowen and Gao, Chang and Huang, Chengen and Lv, Chenxu and others},
  journal={arXiv preprint arXiv:2505.09388},
  year={2025}
}

@article{lambert2024tulu,
  title={Tulu 3: Pushing frontiers in open language model post-training},
  author={Lambert, Nathan and Morrison, Jacob and Pyatkin, Valentina and Huang, Shengyi and Ivison, Hamish and Brahman, Faeze and Miranda, Lester James V and Liu, Alisa and Dziri, Nouha and Lyu, Shane and others},
  journal={arXiv preprint arXiv:2411.15124},
  year={2024}
}

@article{zeng2025simplerl,
  title={Simplerl-zoo: Investigating and taming zero reinforcement learning for open base models in the wild},
  author={Zeng, Weihao and Huang, Yuzhen and Liu, Qian and Liu, Wei and He, Keqing and Ma, Zejun and He, Junxian},
  journal={arXiv preprint arXiv:2503.18892},
  year={2025}
}

@article{yu2025dapo,
  title={Dapo: An open-source llm reinforcement learning system at scale},
  author={Yu, Qiying and Zhang, Zheng and Zhu, Ruofei and Yuan, Yufeng and Zuo, Xiaochen and Yue, Yu and Dai, Weinan and Fan, Tiantian and Liu, Gaohong and Liu, Lingjun and others},
  journal={arXiv preprint arXiv:2503.14476},
  year={2025}
}

@article{lin2025zebralogic,
  title={Zebralogic: On the scaling limits of llms for logical reasoning},
  author={Lin, Bill Yuchen and Bras, Ronan Le and Richardson, Kyle and Sabharwal, Ashish and Poovendran, Radha and Clark, Peter and Choi, Yejin},
  journal={arXiv preprint arXiv:2502.01100},
  year={2025}
}

@inproceedings{wang2022super,
  title={Super-naturalinstructions: Generalization via declarative instructions on 1600+ nlp tasks},
  author={Wang, Yizhong and Mishra, Swaroop and Alipoormolabashi, Pegah and Kordi, Yeganeh and Mirzaei, Amirreza and Naik, Atharva and Ashok, Arjun and Dhanasekaran, Arut Selvan and Arunkumar, Anjana and Stap, David and others},
  booktitle={Proceedings of the 2022 conference on empirical methods in natural language processing},
  pages={5085--5109},
  year={2022}
}

@misc{huan2025doesmathreasoningimprove,
      title={Does Math Reasoning Improve General LLM Capabilities? Understanding Transferability of LLM Reasoning}, 
      author={Maggie Huan and Yuetai Li and Tuney Zheng and Xiaoyu Xu and Seungone Kim and Minxin Du and Radha Poovendran and Graham Neubig and Xiang Yue},
      year={2025},
      eprint={2507.00432},
      archivePrefix={arXiv},
      primaryClass={cs.AI},
      url={https://arxiv.org/abs/2507.00432}, 
}

@misc{bhaskar2025languagemodelsthinkchat,
      title={Language Models that Think, Chat Better}, 
      author={Adithya Bhaskar and Xi Ye and Danqi Chen},
      year={2025},
      eprint={2509.20357},
      archivePrefix={arXiv},
      primaryClass={cs.CL},
      url={https://arxiv.org/abs/2509.20357}, 
}

@article{griffiths2020understanding,
  title={Understanding human intelligence through human limitations},
  author={Griffiths, Thomas L},
  journal={Trends in Cognitive Sciences},
  volume={24},
  number={11},
  pages={873--883},
  year={2020},
  publisher={Elsevier}
}

@article{hu2025open,
  title={Open-reasoner-zero: An open source approach to scaling up reinforcement learning on the base model},
  author={Hu, Jingcheng and Zhang, Yinmin and Han, Qi and Jiang, Daxin and Zhang, Xiangyu and Shum, Heung-Yeung},
  journal={arXiv preprint arXiv:2503.24290},
  year={2025}
}

@article{chen2025acereason,
  title={Acereason-nemotron: Advancing math and code reasoning through reinforcement learning},
  author={Chen, Yang and Yang, Zhuolin and Liu, Zihan and Lee, Chankyu and Xu, Peng and Shoeybi, Mohammad and Catanzaro, Bryan and Ping, Wei},
  journal={arXiv preprint arXiv:2505.16400},
  year={2025}
}

@article{zhao20251,
  title={1.4 million open-source distilled reasoning dataset to empower large language model training},
  author={Zhao, Han and Wang, Haotian and Peng, Yiping and Zhao, Sitong and Tian, Xiaoyu and Chen, Shuaiting and Ji, Yunjie and Li, Xiangang},
  journal={arXiv preprint arXiv:2503.19633},
  year={2025}
}

@misc{openr1,
    title = {Open R1: A fully open reproduction of DeepSeek-R1},
    url = {https://github.com/huggingface/open-r1},
    author = {{Hugging Face}},
    month = {January},
    year = {2025}
}

@article{wang2024mmlu,
  title={Mmlu-pro: A more robust and challenging multi-task language understanding benchmark},
  author={Wang, Yubo and Ma, Xueguang and Zhang, Ge and Ni, Yuansheng and Chandra, Abhranil and Guo, Shiguang and Ren, Weiming and Arulraj, Aaran and He, Xuan and Jiang, Ziyan and others},
  journal={Advances in Neural Information Processing Systems},
  volume={37},
  pages={95266--95290},
  year={2024}
}

@article{guo2025deepseek,
  title={Deepseek-r1: Incentivizing reasoning capability in llms via reinforcement learning},
  author={Guo, Daya and Yang, Dejian and Zhang, Haowei and Song, Junxiao and Wang, Peiyi and Zhu, Qihao and Xu, Runxin and Zhang, Ruoyu and Ma, Shirong and Bi, Xiao and others},
  journal={arXiv preprint arXiv:2501.12948},
  year={2025}
}

@book{newell1972human,
  title={Human problem solving},
  author={Newell, Allen and Simon, Herbert Alexander and others},
  volume={104},
  number={9},
  year={1972},
  publisher={Prentice-hall Englewood Cliffs, NJ}
}

@article{johnson2010mental,
  title={Mental models and human reasoning},
  author={Johnson-Laird, Philip N},
  journal={Proceedings of the National Academy of Sciences},
  volume={107},
  number={43},
  pages={18243--18250},
  year={2010},
  publisher={National Academy of Sciences}
}

@inproceedings{suzgun2023challenging,
  title={Challenging big-bench tasks and whether chain-of-thought can solve them},
  author={Suzgun, Mirac and Scales, Nathan and Sch{\"a}rli, Nathanael and Gehrmann, Sebastian and Tay, Yi and Chung, Hyung Won and Chowdhery, Aakanksha and Le, Quoc and Chi, Ed and Zhou, Denny and others},
  booktitle={Findings of the Association for Computational Linguistics: ACL 2023},
  pages={13003--13051},
  year={2023}
}

@inproceedings{kazemi2025big,
  title={Big-bench extra hard},
  author={Kazemi, Mehran and Fatemi, Bahare and Bansal, Hritik and Palowitch, John and Anastasiou, Chrysovalantis and Mehta, Sanket Vaibhav and Jain, Lalit K and Aglietti, Virginia and Jindal, Disha and Chen, Yuanzhu Peter and others},
  booktitle={Proceedings of the 63rd Annual Meeting of the Association for Computational Linguistics (Volume 1: Long Papers)},
  pages={26473--26501},
  year={2025}
}

@inproceedings{akter2026nemotron,
  title={Nemotron-crossthink: Scaling self-learning beyond math reasoning},
  author={Akter, Syeda Nahida and Prabhumoye, Shrimai and Novikov, Matvei and Han, Seungju and Lin, Ying and Bakhturina, Evelina and Nyberg, Eric and Choi, Yejin and Patwary, Mostofa and Shoeybi, Mohammad and others},
  booktitle={Proceedings of the 19th Conference of the European Chapter of the Association for Computational Linguistics (Volume 1: Long Papers)},
  pages={984--1002},
  year={2026}
}

@article{ma2025general,
  title={General-reasoner: Advancing llm reasoning across all domains},
  author={Ma, Xueguang and Liu, Qian and Jiang, Dongfu and Zhang, Ge and Ma, Zejun and Chen, Wenhu},
  journal={arXiv preprint arXiv:2505.14652},
  year={2025}
}

@article{cheng2026revisiting,
  title={Revisiting reinforcement learning for llm reasoning from a cross-domain perspective},
  author={Cheng, Jorge Zhoujun and Hao, Shibo and Liu, Tianyang and Zhou, Fan and Xie, Yutao and Yao, Feng and Bian, Yuexin and Dey, Nilabjo and Zhuang, Yonghao and Zha, Yuheng and others},
  journal={Advances in Neural Information Processing Systems},
  volume={38},
  year={2026}
}

@article{xia2024less,
  title={Less: Selecting influential data for targeted instruction tuning},
  author={Xia, Mengzhou and Malladi, Sadhika and Gururangan, Suchin and Arora, Sanjeev and Chen, Danqi},
  journal={arXiv preprint arXiv:2402.04333},
  year={2024}
}

@article{guha2025openthoughts,
  title={Openthoughts: Data recipes for reasoning models},
  author={Guha, Etash and Marten, Ryan and Keh, Sedrick and Raoof, Negin and Smyrnis, Georgios and Bansal, Hritik and Nezhurina, Marianna and Mercat, Jean and Vu, Trung and Sprague, Zayne and others},
  journal={arXiv preprint arXiv:2506.04178},
  year={2025}
}

@article{wu2024rose,
  title={Rose: A reward-oriented data selection framework for llm task-specific instruction tuning},
  author={Wu, Yang and Zhang, Huayi and Jiao, Yizheng and Ma, Lin and Liu, Xiaozhong and Yu, Jinhong and Zhang, Dongyu and Yu, Dezhi and Xu, Wei},
  journal={arXiv preprint arXiv:2412.00631},
  year={2024}
}
